\pdfoutput=1

\documentclass[]{interact}

\usepackage{epstopdf}
\usepackage{subfigure}
\usepackage{amsmath,amssymb}
\usepackage{booktabs}
\usepackage{array}
\usepackage[section]{placeins}
\usepackage{algorithm}
\usepackage{algpseudocode}
\usepackage{xurl}%

\usepackage[numbers,sort&compress,merge]{natbib}
\bibpunct[, ]{[}{]}{,}{n}{,}{,}%

\theoremstyle{plain}

\theoremstyle{remark}

\graphicspath{{figures/}}

\newcommand{\thb}{\boldsymbol{\theta}}
\newcommand{\Psib}{\boldsymbol{\Psi}}
\newcommand{\xb}{\mathbf{x}}
\newcommand{\yb}{\mathbf{y}}
\newcommand{\zb}{\mathbf{z}}
\newcommand{\gres}{\mathbf{g}}

\begin{document}

\articletype{ARTICLE}

\title{Sensitivity-Orthogonal Residual Learning for Adaptive Sideslip and
Cornering Stiffness Estimation}

\author{
\name{Kyoungtae Ji\textsuperscript{a}, Dongyeol Won\textsuperscript{b} and Kyoungseok Han\textsuperscript{a}\thanks{CONTACT Kyoungseok Han. Email: kyoungsh@hanyang.ac.kr}}
\affil{\textsuperscript{a}Department of Automotive Engineering, Hanyang University, Seoul, South Korea; \textsuperscript{b}Department of Automotive Engineering (Automotive-Computer Convergence), Hanyang University, Seoul, South Korea}
}

\maketitle
\enlargethispage{4pt}

\begin{abstract}
Online adaptation of the cornering stiffness keeps a model-based sideslip estimator
valid as tires change, but competes with learned model correction for the same
filter innovation. A corrector with enough parameters can absorb the innovation pattern
that a stiffness change produces, and adaptation is suppressed while the fit
improves. We propagate the stiffness sensitivities of the predicted measurements through
the vehicle states and construct a corrector orthogonal to them. A null-space parameterization enforces the constraint without a penalty
coefficient and preserves the static least-squares stiffness estimate on each
identification segment. In full-vehicle simulation, the stiffness response of the
unconstrained corrector declines as its parameter count grows, whereas the orthogonal
corrector responds at least as fully as the uncorrected filter at all sizes and lowers
sideslip error by $12.7\%$ to $22.0\%$. In a simulated drive with a stiffness step, the
orthogonal corrector adapts as quickly as the uncorrected filter and lowers sideslip
error during and after the transient. On public real-vehicle data, the 430-parameter
unconstrained corrector loses most of its response to deliberate offsets of the filter's
nominal stiffness, whereas the orthogonal corrector of the same size retains most of it, with a sideslip
error slightly above the uncorrected filter's.
\end{abstract}


\begin{center}
\small Word count: 6,700 (excluding tables, captions, and references)
\end{center}

\section{Introduction}\label{sec:intro}

Vehicle sideslip angle is a key input to stability control, yet it is not measured in
production vehicles \citep{chindamo2018vehicle}, and the axle cornering stiffnesses that govern its dynamics drift
in service with tire wear, inflation pressure, load, and temperature \citep{pacejka2005tire}. Joint estimation
addresses both problems at once. A model-based filter estimates the sideslip angle
while adapting the stiffness parameters online from the yaw rate and lateral
acceleration. The adapted stiffness has diagnostic value of its own and keeps the
sideslip estimate consistent as the tires change. Holding it constant instead exposes the sideslip
estimate to stiffness mismatch, which raised the error by an order of magnitude in a
simulation study \citep{reina2019augmented}.

The single-track model at the core of such filters is kept simple to limit computation and
the number of parameters to be identified, and its discrepancy from the real vehicle is
systematic rather than random \citep{mosconi2023adaptive}. Methods that combine learned
components with physics-based filters have been proposed to account for this discrepancy
\citep{imbiriba2022apbm,takeishi2023,giampiccolo2024,bertipaglia2024unscented}. These
augmentations reproduce the measurements more closely than the physical model alone. In
the vehicle state estimator closest to ours \citep{bertipaglia2024unscented}, the network
is trained with the stiffnesses fixed at offline-identified values. Its sideslip
estimation error increased when both axle cornering stiffnesses in the model were
increased by $10\%$.

\begin{figure}
\centering
\includegraphics[width=0.95\textwidth]{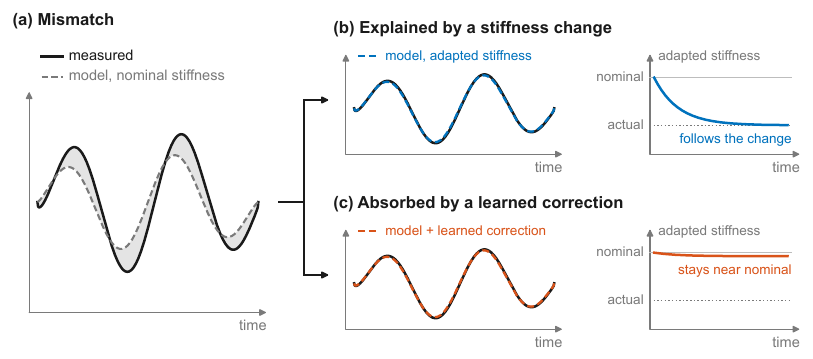}
\caption{One measurement--model mismatch, two explanations, schematic. (a) Measured output
and model prediction at the nominal stiffness. (b) The mismatch removed by adapting the
stiffness, which moves to the actual value. (c) The same mismatch removed by a learned
correction added to the prediction, with the adapted stiffness left near nominal.}
\label{fig:intro}
\end{figure}

As Figure~\ref{fig:intro} illustrates, a mismatch between the measured and the predicted
output can be explained either by a change of the cornering stiffness or by a learned map
of the measured signals, and both explanations reproduce the measurement. A corrector, a learned term added to the model prediction,
absorbs whatever it can predict from its inputs. This includes the component that a
stiffness deviation produces, so the adaptation is suppressed while the fit to the
measurements still improves. This is the confounding between physical parameters and model discrepancy that
is well known in static Bayesian calibration
\citep{kennedy2001bayesian,brynjarsdottir2014learning}. In a sequential filter it appears
as a reduced response of the adapted stiffness to a real stiffness change. The more
parameters the corrector has, the less the filter responds to a real change. The
corrected filter still fits the identification data despite this reduced response, and
the adapted stiffness is rarely compared with an independent reference. Without such a
reference, the adapted values alone do not show whether the response to a stiffness
change is preserved.

The method proposed here constrains the corrector so that its output contains no
component of the form that a stiffness change produces, that is, so that it is orthogonal
to the sensitivities of the predicted outputs to the stiffness. The constraint is enforced
through the parameterization of the corrector. This paper makes three contributions.
\begin{enumerate}
\item We show that the parameter--discrepancy confounding known from static Bayesian
calibration appears in joint sideslip and cornering stiffness estimation as a reduced
response of the adapted stiffness to a real stiffness change.
\item We introduce a residual corrector whose output is orthogonal by construction to
the stiffness sensitivities of the predicted outputs along the state trajectory. The
constraint is built into a null-space parameterization and needs no penalty coefficient.
We show in closed form that the correction leaves the static least-squares stiffness
estimate unchanged.
\item We evaluate stiffness response, axle separation, and transient adaptation together
with sideslip accuracy in full-vehicle simulation. Real-vehicle data with an optical
sideslip reference further test the response to offsets of the nominal stiffness.
\end{enumerate}

\section{Related work}\label{sec:related}

\subsection{Joint sideslip and stiffness estimation}
Cornering stiffness varies in service, and joint state--parameter filters estimate it
online by including the axle stiffnesses in the state vector as random-walk states driven
by production sensors
\citep{bechtoff2016cornering,naets2017twostage,reina2019augmented,viehweger2021guidelines,mosconi2023adaptive,wang2022farckf},
or by estimating them as an adapted noise statistic \citep{berntorp2019tirestiffness}.
Two observations from this literature matter here. First, the
stiffnesses are unobservable in straight driving and only weakly excited at small slip
angles, so the estimates drift unless the adaptation is scheduled by excitation
\citep{naets2017twostage,vanaalst2018adaptive,viehweger2021guidelines} or gated on
observability \citep{berntorp2019tirestiffness}. A batch least-squares fit over a long
window avoids much of the issue \citep{devos2025lsq}. Second, the adapted stiffness is
rarely compared with
an independent reference. The references used are bench tests
\citep{berntorp2019tirestiffness}, offline identification on the same vehicle
\citep{bechtoff2016cornering}, simulation truth \citep{reina2019augmented}, or supplier
values \citep{wang2022farckf}. Where the adapted value is discussed, it
is understood to absorb what the vehicle model omits. The stiffness state is allowed to
track nonlinear and friction-dependent behavior \citep{viehweger2021guidelines} and is noted
to compensate approximations of the suspension kinematics \citep{mosconi2023adaptive}.
None of these adaptive-stiffness estimators is combined with a learned correction term.

\subsection{Learning-augmented filtering}
Augmented physics-based models place a learned component inside a Bayesian filter and
train it jointly with the state estimate \citep{imbiriba2022apbm}. Constraining the
magnitude of the learned contribution has been proposed to preserve the meaning of the
estimated states \citep{straka2025apbm}. Such bounds limit the size of the correction,
whereas the constraint proposed here governs whether the correction resembles a
parameter change. In vehicle state estimation, learned tire models
have been placed inside Kalman filters in place of the stochastic stiffness state, as
offline-trained networks \citep{viehweger2021guidelines} or Gaussian processes
\citep{goblirsch2026gptire}. The architecture closest to ours is a network that
supplies a sideslip pseudo-measurement to a single-track filter and is trained through
it \citep{bertipaglia2024unscented}. The stiffnesses stay at their offline values during
training and operation, and the reported accuracy of the hybrid remains sensitive to a
$\pm 10\%$ stiffness mismatch. Another design instead lets a network generate
time-varying stiffness scaling factors for a bicycle model \citep{kim2024attention}. The
stiffness is then determined entirely by the learned component and, as noted in that
work, cannot be compared with any reference measurement. These designs do not combine a separate
innovation-driven stiffness adaptation with an additive learned corrector.

\subsection{Orthogonal discrepancy and decorrelation}
In static Bayesian calibration, an unconstrained discrepancy term absorbs the
parameter's effect and the calibrated parameter loses its physical meaning
\citep{kennedy2001bayesian,brynjarsdottir2014learning}. Discrepancy priors orthogonal
to the parameter gradients reduce this confounding between the physical parameters and
the discrepancy term \citep{plumlee2017orthogonal}. Soft penalties serve the same purpose
in learning-based modeling. They decorrelate a learned residual from the outputs of the
physics terms \citep{pgnn_orthogonal_ff} or from the parameter Jacobians of the physical
model, as applied to vehicle dynamics in \citep{gyorok2025orthogonal}. The choice of such a regularizer determines which physical
parameters are inferred \citep{takeishi2023}, and mechanistic--neural
non-identifiability can be diagnosed from joint output sensitivities
\citep{giampiccolo2024}. These penalties act on the training loss only. A method for
input--output modeling projects the learned component onto the complement of the
physical regressors' column space by construction, using measured regressors offline
\citep{gyorok2026orthogonal}. This construction also has a kernel-form counterpart
\citep{manna2026}. Neither involves a filter or online parameter adaptation. The present
corrector differs in two respects. It is orthogonal to sensitivities propagated through
unmeasured states, not to measured regressors. It also acts on the innovation that drives
online adaptation in the filter, and on the data used for identification it leaves the
part of that innovation aligned with the sensitivities unchanged.

\section{Joint estimation and its confounding}\label{sec:problem}

\begin{figure}[!htb]
\centering
\includegraphics[width=0.95\textwidth]{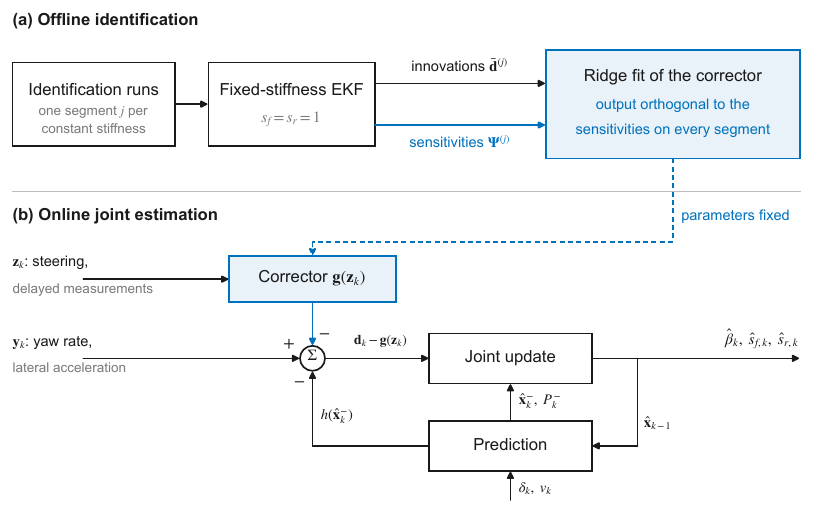}
\caption{Overview of the method. (a) Offline identification of the corrector on segments
of constant stiffness. (b) Online joint estimation with the corrector fixed. Blue: the
elements added by the proposed method. The symbols are defined in
Sections~\ref{sec:problem} and~\ref{sec:method}.}
\label{fig:method}
\end{figure}

Figure~\ref{fig:method} gives an overview of the two stages, offline identification of
the corrector and online joint estimation with the corrector fixed. The corrector is
fitted offline to recorded driving data, a step we call identification. A \emph{segment}
is an interval of identification data over which the stiffness is treated as constant.
On each segment, the filter is run with its stiffness scales fixed at their nominal
value and the other states, including the yaw-rate bias, estimated as usual. We call
this the fixed-stiffness filter. The remaining mismatch between measured and predicted
outputs is the fitting target. The same run also gives the sensitivity of the predicted
outputs to the stiffness, that is, the pattern the mismatch would follow if the
stiffness changed. The corrector is fitted so that, on every segment, its output
contains none of that pattern. Online, the corrector output is subtracted from the mismatch
before the joint update of sideslip and stiffness.

\subsection{Single-track model}\label{sec:model}

\begin{figure}[!htb]
\centering
\includegraphics[width=0.72\textwidth]{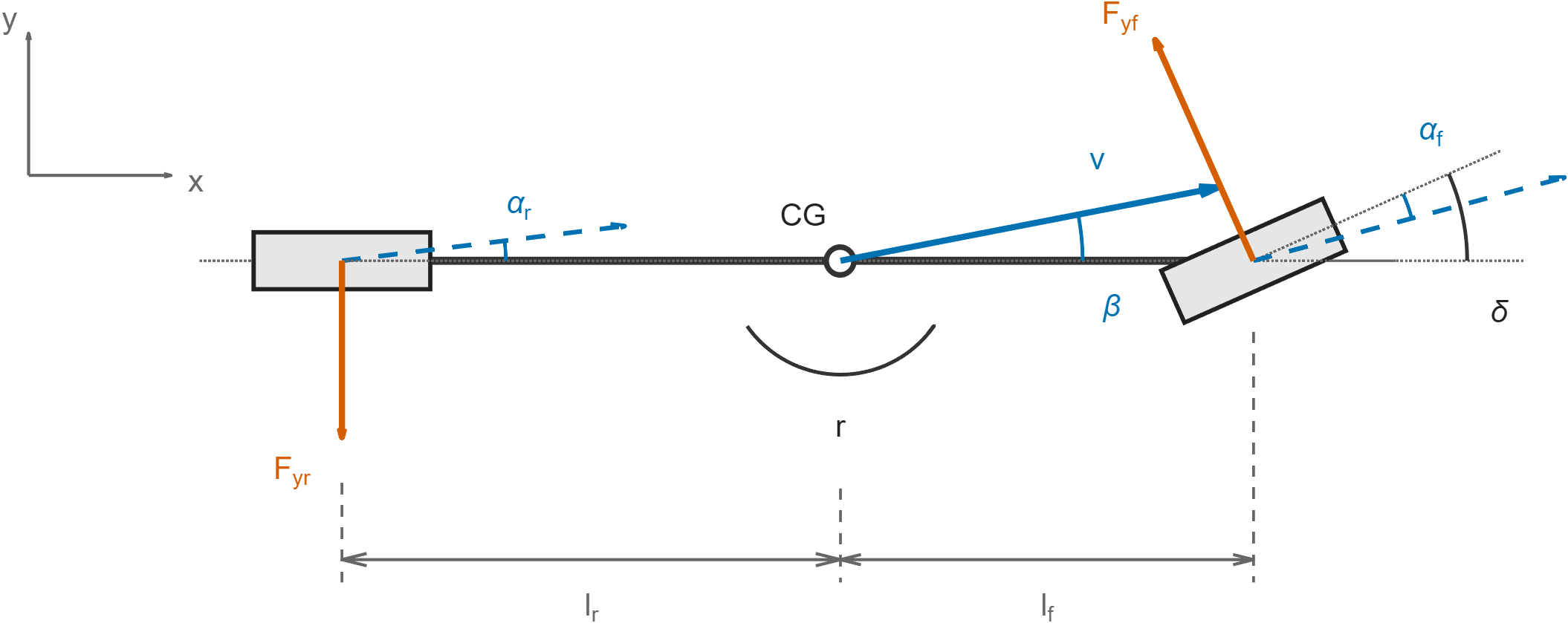}
\caption{Single-track model, top view ($x$ forward, $y$ to the left). Blue: velocities and
slip angles; orange: lateral axle forces.}
\label{fig:model}
\end{figure}

The estimator is built on the single-track model of Pacejka \citep{pacejka2005tire} with
time-varying longitudinal speed, shown in Figure~\ref{fig:model}. The axle slip angles are
\begin{equation}
\alpha_{f,k} = \beta_k + \frac{l_f\,r_k}{v_k} - \delta_k, \qquad
\alpha_{r,k} = \beta_k - \frac{l_r\,r_k}{v_k},
\label{eq:slip_angles}
\end{equation}
where $\beta_k$ is the sideslip angle at the center of gravity, $r_k$ is the yaw rate,
$v_k$ is the measured longitudinal speed, $\delta_k$ is the front-axle steering angle,
and $l_f$ and $l_r$ are the distances from the center of gravity to the front and rear
axles. The axle lateral forces are
\begin{equation}
F_{yf,k} = -s_{f}\,C_{f0}\,\alpha_{f,k}, \qquad
F_{yr,k} = -s_{r}\,C_{r0}\,\alpha_{r,k},
\label{eq:axle_forces}
\end{equation}
where $C_{f0}$ and $C_{r0}$ are fixed nominal axle cornering stiffnesses and $s_f$,
$s_r$ are dimensionless scales adapted online, so that $s_f C_{f0}$ and $s_r C_{r0}$
are the stiffnesses the filter currently uses. The discrete-time dynamics follow from
explicit Euler integration with step $\Delta t$, neglecting the term $-\beta\,\dot v/v$
that a change of speed adds to the sideslip rate:
\begin{equation}
\beta_{k+1} = \beta_k + \Delta t\left(\frac{F_{yf,k}+F_{yr,k}}{m\,v_k} - r_k\right),
\qquad
r_{k+1} = r_k + \Delta t\,\frac{l_f\,F_{yf,k}-l_r\,F_{yr,k}}{I_z},
\label{eq:dynamics}
\end{equation}
where the first term is the lateral acceleration $a_{y,k} = (F_{yf,k}+F_{yr,k})/m$ divided
by the speed, $m$ is the vehicle mass, and $I_z$ is the yaw moment of inertia.

The measurement vector contains the yaw rate and the lateral acceleration,
\begin{equation}
\yb_k = h(\xb_k) + \mathbf{e}_k, \qquad
h(\xb_k) =
\begin{bmatrix} r_k + b_{r,k} \\[2pt] a_{y,k} \end{bmatrix},
\label{eq:measurement}
\end{equation}
where $\xb_k = [\beta_k,\, r_k,\, s_f,\, s_r,\, b_{r,k}]^{\top}$ collects the dynamic
states, the stiffness scales, and a slowly varying bias $b_{r,k}$ of the yaw-rate
measurement. The two components of $\yb_k$ are the measured yaw rate $r^{\mathrm{m}}_k$
and the measured lateral acceleration $a^{\mathrm{m}}_{y,k}$. The two components of $h$
are their noise-free predictions: the yaw-rate state with the bias added, and the
lateral acceleration $a_{y,k}$. $\mathbf{e}_k$ is zero-mean measurement
noise with covariance $R = \operatorname{diag}(\sigma_r^2,\ \sigma_{a_y}^2)$, where
$\sigma_r$ and $\sigma_{a_y}$ are the standard deviations of the two measurement noises.

\subsection{Joint extended Kalman filter}\label{sec:ekf}

Joint estimation is performed by an extended Kalman filter (EKF) on the augmented state
$\xb_k$. The stiffness scales, collected in $\thb=[s_f,\, s_r]^{\top}$, and the yaw-rate
bias are modeled as bounded random walks. For example, the front stiffness scale follows
$s_{f,k+1} = s_{f,k} + w_k$ with $w_k \sim \mathcal{N}(0, q)$, allowing small changes at
each step. The noise keeps the corresponding entry of the state covariance from shrinking to
zero, so the stiffness scales remain adjustable. The update moves $\thb$ when the
maneuver excites the stiffness and the innovation contains a component consistent with
a stiffness change. This is the conventional
structure for joint sideslip and stiffness estimation
\citep{bechtoff2016cornering,naets2017twostage,reina2019augmented,viehweger2021guidelines,mosconi2023adaptive}.
Local observability from yaw rate and lateral acceleration under steering excitation has
been shown for a related augmented-state model \citep{reina2019augmented}.

The filter itself is a standard EKF. Let $\mathbf{f}$ denote the augmented transition,
which applies \eqref{eq:dynamics} to the dynamic states and the identity to the
stiffness scales and the bias, and let $F_k=\partial \mathbf{f}/\partial\xb$ be its
Jacobian evaluated at $\hat{\xb}_{k-1}$. Here $\hat{\xb}_k^{-}$ and $\hat{\xb}_k$ are the
state estimates before and after the measurement update, and $P_k^{-}$ and $P_k$ are
their $5\times 5$ error covariances. $Q$ is the diagonal $5\times 5$ process-noise
covariance per sample. The prediction is
\begin{equation}
\hat{\xb}_k^{-} = \mathbf{f}(\hat{\xb}_{k-1}), \qquad
P_k^{-} = F_k P_{k-1} F_k^{\top} + Q .
\label{eq:ekf_predict}
\end{equation}
The predicted measurement is $\hat{\yb}_k = h(\hat{\xb}_k^{-})$, and
$\yb_k - \hat{\yb}_k$ is the innovation. With $H_k=\partial h/\partial\xb$ at the
predicted state, the update is
\begin{equation}
\begin{aligned}
K_k &= P_k^{-}H_k^{\top}\big(H_k P_k^{-} H_k^{\top} + R\big)^{-1}, \\
\hat{\xb}_k &= \hat{\xb}_k^{-} + K_k\,(\yb_k - \hat{\yb}_k), \\
P_k &= (I - K_k H_k)\,P_k^{-}.
\end{aligned}
\label{eq:ekf_update}
\end{equation}
$K_k$ denotes the $5\times 2$ Kalman gain, and $I$ denotes the identity matrix.
The stiffness scales are kept within a physical range by a covariance-weighted
projection of the joint state estimate onto the bounds, with the corresponding
covariance update \citep{simon2010kalman}. At low speed, the stiffness scales are not
updated. The EKF with the stiffness scales adapted online and no corrector is the
uncorrected filter.

\subsection{Residual corrector and innovation competition}\label{sec:stealing}

A single-track model with fixed parameters cannot represent load transfer, steering
compliance, or other unmodeled dynamics of a real vehicle. A standard approach is to add
a learned residual correction to the predicted measurement. This corrector is
\begin{equation}
\gres(\zb_k) = W^{\top}\phi(\zb_k),
\label{eq:corrector}
\end{equation}
where $\zb_k$ collects the signals available at time $k$, $\phi(\zb_k)$ is a vector of
$n_\phi$ regressors computed from these signals, and $W$ is an $n_\phi\times 2$
parameter matrix identified from data. The corrector is linear in its parameters $W$.
The construction below applies to any linear output layer whose regressors are not
trained, including the last layer of a network. The regressors are strictly
causal, so that the corrector cannot use the current measurement to reproduce it.
Measured signals enter with a delay of at least one sample, and only the steering input,
which is a driver command and not a measured response, enters at the current time.

The corrector output is added to the predicted measurement of the filter,
\begin{equation}
\hat{\yb}_k = h(\hat{\xb}_k^{-}) + \gres(\zb_k).
\label{eq:ekf_meas}
\end{equation}
The corrector input $\zb_k$ contains only measured signals and the steering input, not
the state, so $H_k$ acquires no term from the correction. The correction changes the gain only
through the state trajectory, the sequence of state estimates along which $F_k$ and
$H_k$ are evaluated. With $\gres\equiv 0$ the filter is the uncorrected filter.

Let $\xb^{s}_k=[\beta_k, r_k]^{\top}$ denote the dynamic states and
$\mathbf{d}_k = \yb_k - h(\hat{\xb}_k^{-})$ the innovation the filter would have without
the corrector, so that $\yb_k - \hat{\yb}_k = \mathbf{d}_k - \gres(\zb_k)$. Let
$K_{x,k}$ and $K_{\theta,k}$ denote the $2\times 2$ row blocks of $K_k$ corresponding to
the dynamic states and the stiffness scales, respectively. With the bias row omitted,
the update in \eqref{eq:ekf_update} is
\begin{equation}
\begin{bmatrix} \hat{\xb}^{s}_k \\[2pt] \hat{\thb}_k \end{bmatrix}
=
\begin{bmatrix} \hat{\xb}^{s-}_k \\[2pt] \hat{\thb}^{-}_k \end{bmatrix}
+
\begin{bmatrix} K_{x,k} \\[2pt] K_{\theta,k} \end{bmatrix}
\big(\mathbf{d}_k - \gres(\zb_k)\big).
\label{eq:ekf_blocks}
\end{equation}
The stiffness rows $K_{\theta,k}$ act on $\mathbf{d}_k - \gres(\zb_k)$, not on
$\mathbf{d}_k$. The corrector and the stiffness adaptation compete for the same
innovation. This innovation competition is the form that the parameter--discrepancy
confounding takes in the filter, and its result is a suppressed stiffness adaptation. If
the corrector output reproduces the part of $\mathbf{d}_k$ that a stiffness change
produces, that part is subtracted before the stiffness scales are updated. The absorbed
part need not come from a stiffness deviation. A component of the identification
innovation that is merely correlated with the innovation a stiffness deviation would
produce is absorbed as well, and this can reduce the response of the adapted stiffness
to a real change. When the stiffness later differs from its value during
identification, the suppressed adaptation compensates less of the difference, and the
sideslip estimate can degrade. The corrector is identified on the fixed-stiffness filter
innovations because they retain both the model discrepancy and the stiffness mismatch
before stiffness adaptation absorbs either.

\subsection{Trajectory sensitivity to the stiffness scales}\label{sec:sensitivity}

To measure the confounding, we first derive the change in the predicted outputs that a
small stiffness change would produce over a segment. Run the fixed-stiffness model along
the steering and speed of one segment once at the nominal $s_f=s_r=1$ and once with
either the front or the rear scale raised to $1+\Delta s$. The difference of the two
predicted $[r, a_y]$ sequences, divided by $\Delta s$, is the sensitivity of the predicted
outputs to that scale, and multiplied by a small change of the scale it gives the
innovation produced by that change. In practice the model is not run twice. We compute
the derivative that this difference approximates by propagating the linearized model
along the state trajectory of the fixed-stiffness filter, hereafter the nominal
trajectory, as in \eqref{eq:sensitivity} below. The linearization point is therefore the
filter's state estimate, not the state of a model run on its own. The constraint and the
static estimate below both use this sensitivity.

A stiffness change affects the measurements in two ways: directly, through the axle
forces, and indirectly, through the sideslip and yaw-rate response that follows. The
indirect effect is essential because sideslip is not measured and its change reaches the
measurements only through the dynamics. In straight driving without steering excitation
the axle slip angles are zero, so the nominal axle forces and with them the stiffness
sensitivities vanish.

The transition \eqref{eq:dynamics} and the measurement map $h$ are linearized with
respect to the dynamic states $\xb^{s}$ and the stiffness scales $\thb$ along the
nominal trajectory at $\thb_0=[1,1]^{\top}$. This gives four Jacobians: $A_k$ and $B_k$
of the next dynamic state with respect to the current dynamic state and to the stiffness
scales, and $C_k$ and $D_k$ of the measurement map with respect to the same two,
\begin{equation}
\begin{aligned}
A_k &=
\begin{bmatrix} 1 & -\Delta t \\[2pt] 0 & 1 \end{bmatrix}
+ \Delta t
\begin{bmatrix}
(\mathbf{c}_{f,k}+\mathbf{c}_{r,k})/(m v_k) \\[2pt]
(l_f\,\mathbf{c}_{f,k}-l_r\,\mathbf{c}_{r,k})/I_z
\end{bmatrix},
&\qquad C_k &=
\begin{bmatrix}
0 & 1 \\[2pt]
(\mathbf{c}_{f,k}+\mathbf{c}_{r,k})/m
\end{bmatrix},\\[4pt]
B_k &= \Delta t
\begin{bmatrix}
\mathbf{F}^{0}_k/(m v_k) \\[2pt]
[\,l_f\,F^{0}_{yf,k},\; -l_r\,F^{0}_{yr,k}\,]/I_z
\end{bmatrix},
&\qquad D_k &=
\begin{bmatrix}
0 & 0 \\[2pt]
\mathbf{F}^{0}_k/m
\end{bmatrix},
\end{aligned}
\label{eq:jacobians}
\end{equation}
where $\mathbf{c}_{f,k} = -C_{f0}\,[1,\; l_f/v_k]$ and
$\mathbf{c}_{r,k} = -C_{r0}\,[1,\; -l_r/v_k]$ are the gradients of the axle forces with
respect to the dynamic states, and $\mathbf{F}^{0}_k = [\,F^{0}_{yf,k},\; F^{0}_{yr,k}\,]$
collects the nominal axle forces along the trajectory,
$F^{0}_{yf,k}=-C_{f0}\alpha_{f,k}$ and $F^{0}_{yr,k}=-C_{r0}\alpha_{r,k}$. The first
row of $D_k$ is zero because a stiffness change reaches the yaw rate only through the
states. These four matrices have the same form as the blocks of $F_{k+1}$ and $H_k$ in
\eqref{eq:ekf_predict} and \eqref{eq:ekf_update} with respect to the dynamic states and
the stiffness scales. The filter evaluates those blocks along its own state estimates,
and the sensitivity recursion evaluates them along the nominal trajectory. The state and
output sensitivities to the stiffness scales are propagated along the segment,
\begin{equation}
S_{k+1} = A_k S_k + B_k, \qquad
G_k = C_k S_k + D_k, \qquad S_1 = 0.
\label{eq:sensitivity}
\end{equation}
Here $S_k \in \mathbb{R}^{2\times 2}$ is the sensitivity of the dynamic states at time
$k$ to a stiffness perturbation kept constant since the start of the segment. A
perturbation $\delta\thb$ moves the state by $S_k\delta\thb$ to first order.
$G_k \in \mathbb{R}^{2\times 2}$ is the corresponding sensitivity of the predicted
outputs. $S_1=0$ because the state has not yet responded at the start of the segment.
Each $G_k$ is multiplied by $R^{-1/2} = \operatorname{diag}(1/\sigma_r,\ 1/\sigma_{a_y})$,
so that the yaw rate and the lateral acceleration enter the estimate \eqref{eq:readout}
and the constraint \eqref{eq:orth_meaning} in units of their measurement-noise standard
deviations. Stacking the weighted
$G_k$ over a segment of $N_j$
samples gives the sensitivity matrix
\begin{equation}
\Psib^{(j)} =
\begin{bmatrix} R^{-1/2} G_1 \\ \vdots \\ R^{-1/2} G_{N_j} \end{bmatrix}
= [\,\boldsymbol{\psi}_f,\ \boldsymbol{\psi}_r\,] \in \mathbb{R}^{2N_j \times 2},
\label{eq:psi_stack}
\end{equation}
where $j$ indexes the segment and is omitted when one segment is discussed. The columns
$\boldsymbol{\psi}_f$ and $\boldsymbol{\psi}_r$ are the weighted output sensitivities to
the front and the rear stiffness scale over the segment. The column space of
$\Psib$, $\operatorname{col}(\Psib)$, describes the output perturbations produced by
constant stiffness changes under this local linearization.

The recursion \eqref{eq:sensitivity} does not differentiate the filter update, because
the quantity to preserve is the perturbation that a parameter change produces in the
predicted measurements. A corrector output aligned with these directions
explains the same output variation as a stiffness change, even when the corrector output
is small. The constraint therefore acts on the projection of the correction onto these
directions.

\subsection{Static estimate of the stiffness change from the innovation}\label{sec:lsview}

The sensitivity matrix gives a static estimate of the stiffness change contained in an
innovation sequence, computed over the whole segment at once rather than sample by sample
as in the filter. As in the batch least-squares identification of
\citep{devos2025lsq}, we regress the stacked innovation on the two sensitivity columns:
\begin{equation}
\widehat{\delta\thb}(\bar{\mathbf{d}}) =
\left(\Psib^{\top}\Psib\right)^{-1}\Psib^{\top}\bar{\mathbf{d}},
\label{eq:readout}
\end{equation}
where $\Psib$ has full column rank and $\bar{\mathbf{d}} \in \mathbb{R}^{2N_j}$ is the
$R^{-1/2}$-weighted stack of fixed-stiffness filter innovations on the segment, so that
\eqref{eq:readout} gives the coefficients of the projection of $\bar{\mathbf{d}}$ onto
the columns of $\Psib$. The projection itself,
$\Psib(\Psib^{\top}\Psib)^{-1}\Psib^{\top}\bar{\mathbf{d}}$, is the sensitivity-aligned
component of the innovation. Let the innovation be the sum of the effect
$\Psib\,\delta\thb^{\ast}$ of a true stiffness deviation $\delta\thb^{\ast}$, a model
discrepancy term $\bar{\mathbf{m}}$, and zero-mean noise $\bar{\mathbf{e}}$,
\begin{equation}
\bar{\mathbf{d}} = \Psib\,\delta\thb^{\ast} + \bar{\mathbf{m}} + \bar{\mathbf{e}}.
\label{eq:innov_decomp}
\end{equation}
With the sensitivity matrix treated as fixed, the expectation of \eqref{eq:readout} is
\begin{equation}
\mathbb{E}[\widehat{\delta\thb}] = \delta\thb^{\ast}
+ \left(\Psib^{\top}\Psib\right)^{-1}\Psib^{\top}\bar{\mathbf{m}}.
\label{eq:readout_decomp}
\end{equation}
The second term is the contribution of the model discrepancy to the estimated stiffness
change \eqref{eq:readout}. Because of this term, the static estimate
of a single-track filter is an effective model stiffness and not the physical cornering
stiffness of the tires.

An unconstrained corrector, with output $\bar{\gres}$ on the segment stacked and weighted
like $\bar{\mathbf{d}}$, is fitted to minimize the remaining innovation
$\bar{\mathbf{d}} - \bar{\gres}$. It absorbs the part of $\bar{\mathbf{d}}$ that its
regressors can express, including any component along the columns of $\Psib$, whether
from a true stiffness deviation $\Psib\,\delta\thb^{\ast}$ or from the discrepancy
$\bar{\mathbf{m}}$. The estimate \eqref{eq:readout} on the remaining innovation then
shifts by $-(\Psib^{\top}\Psib)^{-1}\Psib^{\top}\bar{\gres}$, and a corrector that
reproduces $\Psib\,\delta\thb^{\ast}$ removes the stiffness deviation from it.

\section{Sensitivity-orthogonal residual learning}\label{sec:method}

\subsection{Orthogonality condition}\label{sec:orth_meaning}

\begin{figure}[!htb]
\centering
\includegraphics[width=0.95\textwidth]{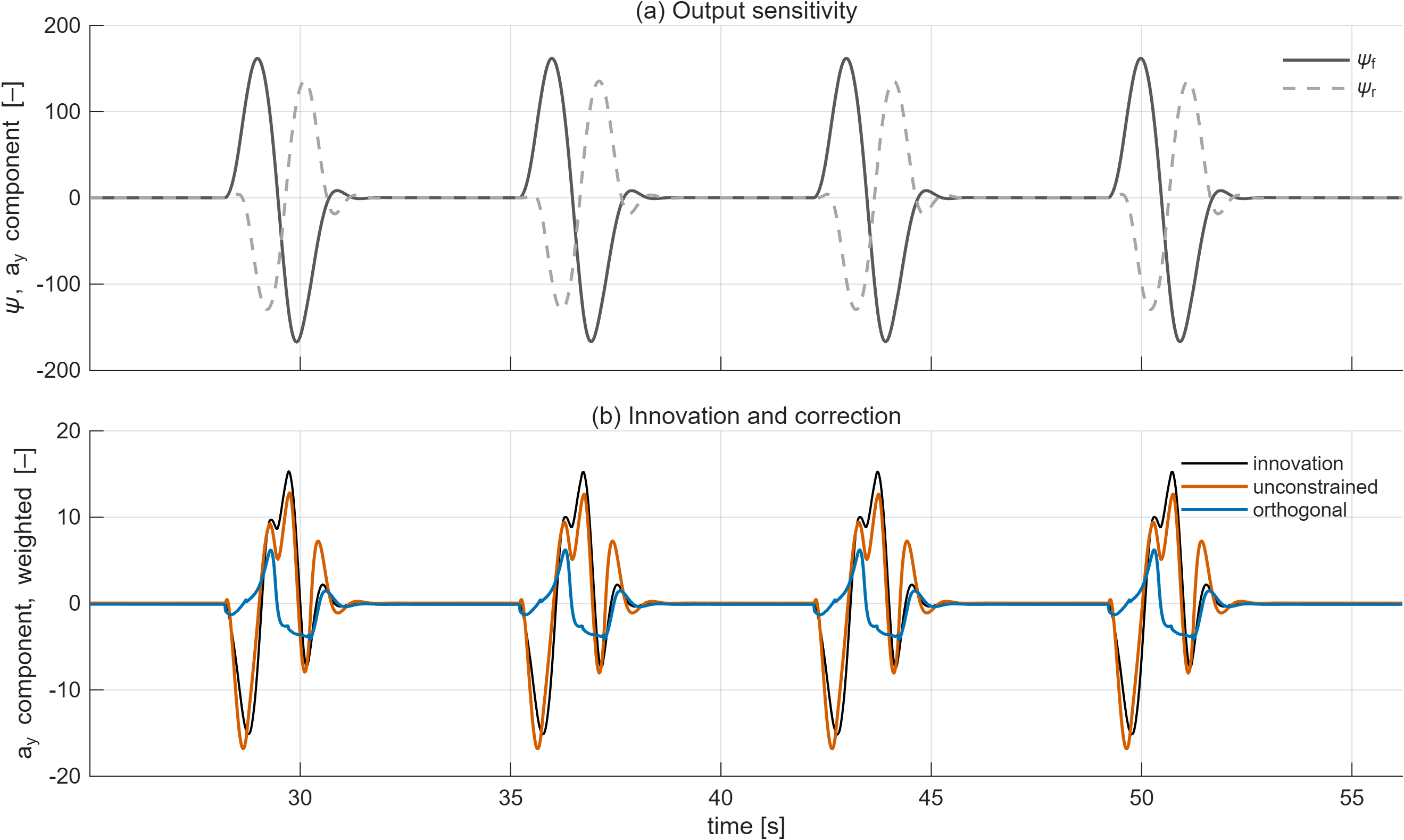}
\caption{Identification run with sinusoidal steering at stiffness scale $0.8$,
lateral-acceleration component, $R^{-1/2}$-weighted. (a) Sensitivities
$\boldsymbol{\psi}_f$ (solid) and $\boldsymbol{\psi}_r$ (dashed). (b) Fixed-stiffness
filter innovation and the outputs of the unconstrained and the orthogonal corrector.
Both correctors have 30 parameters.}
\label{fig:sens}
\end{figure}

Let $\bar{\gres}^{(j)} \in \mathbb{R}^{2N_j}$ be the corrector output on segment $j$,
stacked and weighted like $\Psib^{(j)}$. The proposed constraint is that its inner
product with each sensitivity column is zero,
\begin{equation}
\boldsymbol{\psi}_f^{\top}\,\bar{\gres}^{(j)} = 0, \qquad
\boldsymbol{\psi}_r^{\top}\,\bar{\gres}^{(j)} = 0,
\qquad\text{i.e.}\qquad \Psib^{(j)\top}\bar{\gres}^{(j)} = \mathbf{0},
\label{eq:orth_meaning}
\end{equation}
so that no part of the corrector output on the segment has the form of a front or a rear
stiffness change. Figure~\ref{fig:sens} shows the sensitivities and the outputs of the
two correctors on an identification run at stiffness scale $0.8$ from the simulation
study of Section~\ref{sec:sim}. The unconstrained output follows the main waveform of
the lateral-acceleration innovation. Evaluated, as the condition requires, over both
measured signals and the whole segment, it also absorbs most of the sensitivity-aligned
component, whereas the orthogonal correction leaves this component in the remaining
innovation.

\begin{figure}[t]
\centering
\includegraphics[width=0.95\textwidth]{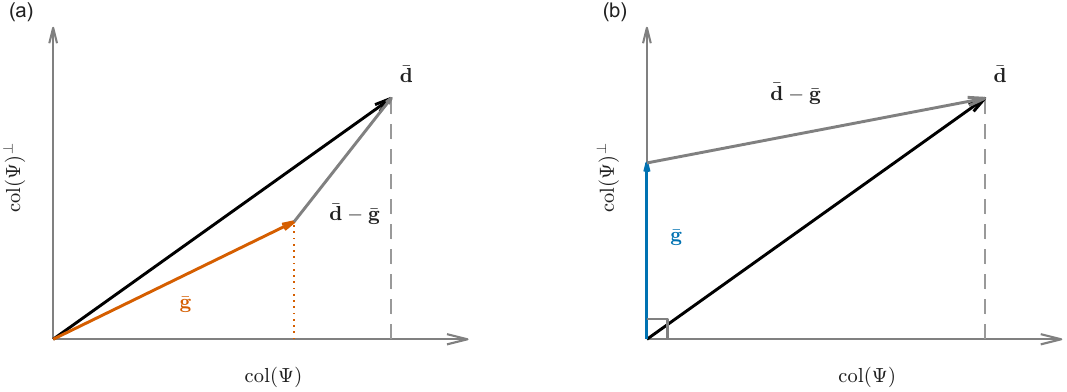}
\caption{Geometric view of the orthogonality constraint, schematic, in the plane spanned
by $\operatorname{col}(\Psib)$ and its orthogonal complement. (a) Unconstrained
correction. (b) Sensitivity-orthogonal correction.}
\label{fig:concept}
\end{figure}

Applying the static estimate \eqref{eq:readout} to the remaining innovation gives
\begin{equation}
\begin{aligned}
\widehat{\delta\thb}\big(\bar{\mathbf{d}} - \bar{\gres}\big)
&= \left(\Psib^{\top}\Psib\right)^{-1}\Psib^{\top}\bar{\mathbf{d}}
- \left(\Psib^{\top}\Psib\right)^{-1}\Psib^{\top}\bar{\gres} \\
&= \widehat{\delta\thb}\big(\bar{\mathbf{d}}\big),
\end{aligned}
\label{eq:invariance}
\end{equation}
because $\Psib^{\top}\bar{\gres} = \mathbf{0}$ by \eqref{eq:orth_meaning}. The static
estimate of the stiffness change is therefore the same before and after correction.
In Figure~\ref{fig:concept}, an unconstrained correction has a component along
$\operatorname{col}(\Psib)$ and changes the projection of the remaining innovation onto
the sensitivities (a), whereas a sensitivity-orthogonal correction lies in the
complement and leaves that projection unchanged (b).

The identity holds for the static estimate on every segment on which
\eqref{eq:orth_meaning} is enforced. The sequential filter does not average over a whole
segment. It weights recent innovations through its gains, and the correction also
changes the estimated state trajectory along which those gains are computed. We
therefore measure experimentally how much of the filter's response to a stiffness
change is preserved, in Section~\ref{sec:sim} by the response of the adapted stiffness
to a known change and in Section~\ref{sec:real} by its response to an offset of the
nominal stiffness.

\subsection{Constrained identification}\label{sec:fit}

The corrector output is linear in its parameters. Let $\mathbf{w} \in
\mathbb{R}^{2 n_\phi}$ denote the stacked columns of $W$ and $\bar{\Phi}^{(j)} \in
\mathbb{R}^{2N_j \times 2n_\phi}$ the regressor matrix that maps $\mathbf{w}$ to the
weighted, stacked output on segment $j$, $\bar{\gres}^{(j)} = \bar{\Phi}^{(j)}\mathbf{w}$. The constraint \eqref{eq:orth_meaning}
on segment $j$ is then the pair of linear equations
$\Psib^{(j)\top}\bar{\Phi}^{(j)}\mathbf{w} = \mathbf{0}$. The constraints of all $J$
segments are collected into
\begin{equation}
\mathcal{C}\,\mathbf{w} = \mathbf{0}, \qquad
\mathcal{C} =
\begin{bmatrix}
\Psib^{(1)\top}\bar{\Phi}^{(1)} \\ \vdots \\ \Psib^{(J)\top}\bar{\Phi}^{(J)}
\end{bmatrix} \in \mathbb{R}^{2 J \times 2 n_\phi}.
\label{eq:constraint_matrix}
\end{equation}
The constraints are imposed per segment because nonzero inner products between the
correction and the stiffness sensitivities can cancel across segments in a pooled
constraint.

Let $\bar{\mathbf{d}} = [\bar{\mathbf{d}}^{(1)}; \dots; \bar{\mathbf{d}}^{(J)}] \in
\mathbb{R}^{2N}$ and $\bar{\Phi} = [\bar{\Phi}^{(1)}; \dots; \bar{\Phi}^{(J)}] \in
\mathbb{R}^{2N \times 2n_\phi}$ stack the weighted innovations and regressors of all
segments, with $N = \sum_j N_j$. The nonconstant regressors listed in
Table~\ref{tab:settings} are standardized to zero mean and unit variance on the
identification data before the fit. The constant regressor remains equal to one. The
identification is a ridge regression of the fixed-stiffness filter innovations on the
regressors, restricted to the parameters that satisfy \eqref{eq:constraint_matrix}:
\begin{equation}
\min_{\mathbf{w}}\;
\left\lVert \bar{\Phi}\,\mathbf{w} - \bar{\mathbf{d}} \right\rVert^{2}
+ \rho\,N \lVert \mathbf{w} \rVert^{2}
\qquad \text{s.t.} \qquad \mathcal{C}\,\mathbf{w} = \mathbf{0},
\label{eq:constrained_fit}
\end{equation}
where $\rho>0$ is a ridge coefficient, scaled by $N$ so that it does not depend on the
amount of identification data. Let the columns of
$Z \in \mathbb{R}^{2 n_\phi \times (2 n_\phi - \operatorname{rank}\mathcal{C})}$ be an
orthonormal basis of the null space of $\mathcal{C}$, so that every
$\mathbf{w} = Z\boldsymbol{\eta}$ with a reduced parameter vector
$\boldsymbol{\eta} \in \mathbb{R}^{2 n_\phi - \operatorname{rank}\mathcal{C}}$ satisfies
the constraint identically. The ridge problem in the reduced coordinates has the closed-form
solution
\begin{equation}
\boldsymbol{\eta} =
\left(Z^{\top}\bar{\Phi}^{\top}\bar{\Phi}\,Z + \rho N I\right)^{-1}
Z^{\top}\bar{\Phi}^{\top}\bar{\mathbf{d}},
\qquad
\mathbf{w} = Z\,\boldsymbol{\eta}.
\label{eq:nullspace}
\end{equation}
The constraint thus holds exactly on every identification segment, without a penalty
coefficient for it, unlike the soft penalties of
\citep{pgnn_orthogonal_ff,gyorok2025orthogonal}. The construction removes
$\operatorname{rank}\mathcal{C}$ parameter dimensions, so
$2n_\phi - \operatorname{rank}\mathcal{C}$ parameters remain free. With independent
constraints $\operatorname{rank}\mathcal{C} = 2J$, and the corrector then needs more
than $J$ regressors to retain any free parameters. Finer segmentation gives a stricter guarantee
but leaves fewer free parameters, so each identification run is one segment, the
longest interval over which the stiffness is constant. Algorithm~\ref{alg:main} summarizes the complete procedure.

\begin{algorithm}[t]
\caption{Identification and online operation of the orthogonal corrector.}
\label{alg:main}
\begin{algorithmic}[1]
\Statex \textbf{Offline identification} (on $J$ segments of constant stiffness)
\State Run the fixed-stiffness EKF ($s_f=s_r=1$) on each segment to obtain its innovations and state trajectory.
\State Form the $R^{-1/2}$-weighted regressor matrices $\bar{\Phi}^{(j)}$ and innovation vectors $\bar{\mathbf{d}}^{(j)}$.
\State Compute the sensitivity matrices $\Psib^{(j)}$ along these trajectories using \eqref{eq:sensitivity} and~\eqref{eq:psi_stack}.
\State Form the constraint matrix $\mathcal{C}$ using \eqref{eq:constraint_matrix} and compute an orthonormal basis $Z$ of its null space.
\State Determine the corrector parameters $W$ using the null-space ridge solution \eqref{eq:nullspace}.
\Statex \vspace{-0.6\baselineskip}\hspace*{-\algorithmicindent}\rule{\dimexpr\linewidth+\algorithmicindent\relax}{0.4pt}
\Statex \textbf{Online operation} (every sample, $W$ fixed)
\State Predict the state and covariance using \eqref{eq:ekf_predict}.
\State Add the corrector output $\gres(\zb_k)$ to the predicted measurement using \eqref{eq:ekf_meas}.
\State Update the state and covariance using \eqref{eq:ekf_update}.
\end{algorithmic}
\end{algorithm}

\section{Controlled simulation study}\label{sec:sim}

All simulation experiments use the full-vehicle demonstration passenger-car model in IPG
CarMaker. Its tire cornering stiffness is scaled through the cornering-stiffness scaling
factor $\lambda_{Ky}$ of its Magic Formula tire model \citep{pacejka2005tire}.
The estimator uses only yaw rate, lateral acceleration, vehicle speed, and the
road-wheel steering angle at $100$~Hz. Vehicle parameters and filter settings are listed
in Table~\ref{tab:settings}, and the filter settings are the same for every method
and experiment in this section.

We evaluate stiffness adaptation against the effective axle stiffness of the
single-track model. This quantity can differ from the tire stiffness because it also
reflects unmodeled effects. For each run, this reference is obtained by regressing the
axle forces on the model slip angles computed from the true vehicle states and the
steering input. The regression conditions are given in Appendix~\ref{app:settings}.

\subsection{Experimental setup and metrics}\label{sec:sim_setup}

\begin{figure}[t]
\centering
\includegraphics[width=0.95\textwidth]{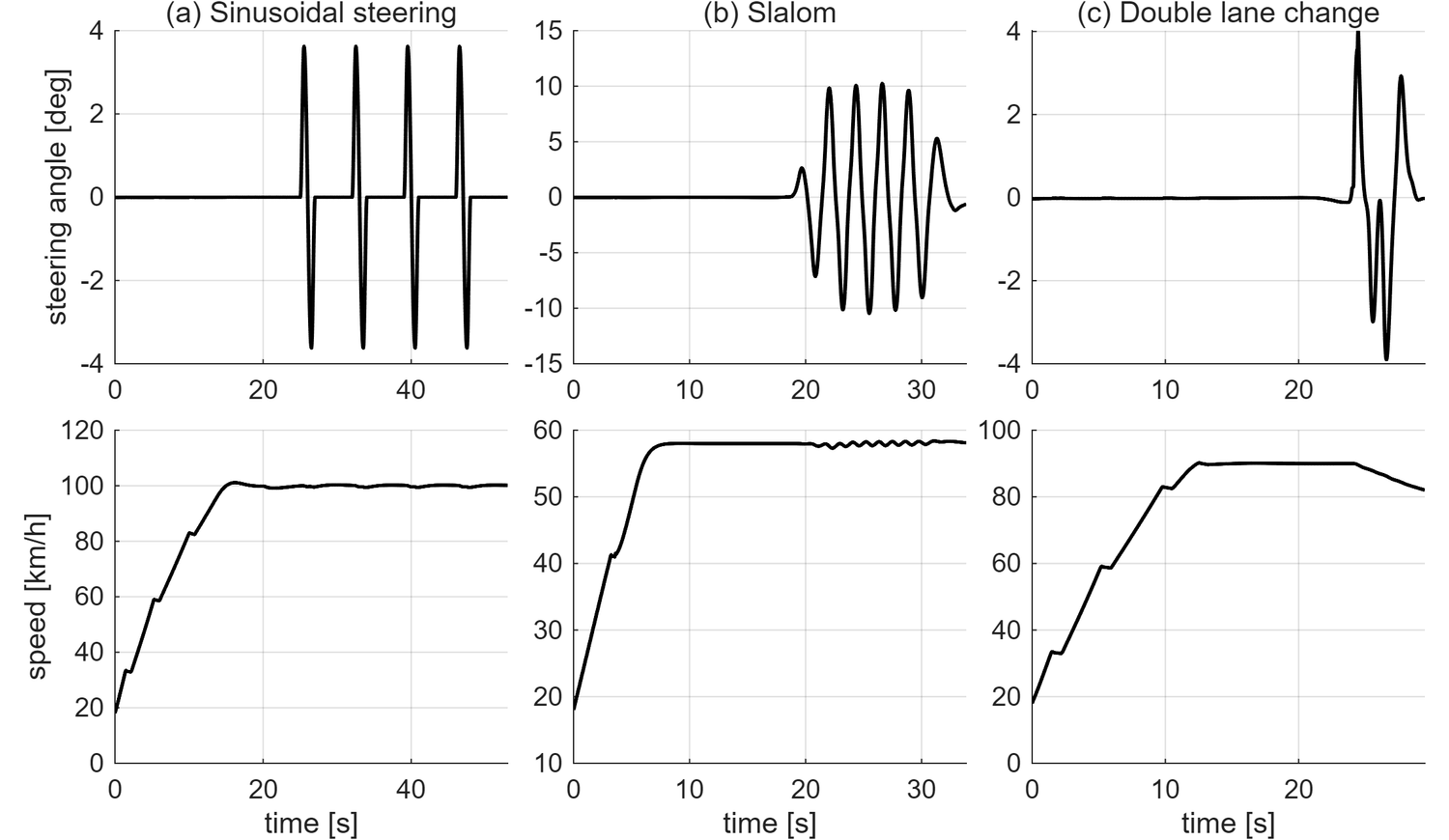}
\caption{Steering angle (top) and speed (bottom) of the identification maneuvers in
simulation at the nominal stiffness. (a) Sinusoidal steering. (b) Slalom. (c) Double
lane change.}
\label{fig:inputs}
\end{figure}

Table~\ref{tab:matrix} lists the runs by role, maneuver, and stiffness factor. The
identification set consists of eleven runs at
constant stiffness: the three maneuvers at three stiffness scales, and two step-steering
drives at nominal stiffness, at $80$ and $50$~km/h, which add a step input and a second
speed to the identification data. The $80$~km/h drive also serves as the constant-stiffness reference
in Section~\ref{sec:continuous}. These two drives use only the nominal stiffness because
each further segment removes two free parameters from the orthogonal corrector. Every
other run is an evaluation run
whose condition is absent from identification. Figure~\ref{fig:inputs} shows the
steering and speed of the three identification maneuvers.

\begin{table}[t]
\centering
\caption{Simulation runs. The stiffness factor $\lambda_{Ky}$ applies to all tires unless stated.}
\label{tab:matrix}
\footnotesize
\setlength{\tabcolsep}{3pt}
\begin{tabular}{@{}>{\raggedright\arraybackslash}p{0.29\textwidth} >{\raggedleft\arraybackslash}p{0.05\textwidth} >{\raggedright\arraybackslash}p{0.42\textwidth} >{\raggedright\arraybackslash}p{0.195\textwidth}@{}}
\toprule
Role & Runs & Maneuver & $\lambda_{Ky}$ \\
\midrule
Identification & 9 & sinusoidal steering, slalom, double lane change & 0.8, 1.0, 1.1 \\
Identification & 2 & step steering at 80 and 50~km/h & 1.0 \\
Evaluation, intermediate scale & 3 & the three maneuvers above & 0.9 \\
Evaluation, one axle & 8 & sinusoidal steering, slalom & 0.85 or 1.15, one axle \\
Evaluation, change during drive & 1 & step steering at 80~km/h, 360~s & 1.0, 0.8 in 120--240~s \\
\bottomrule
\end{tabular}
\end{table}

We compare three estimators, each the EKF of Section~\ref{sec:ekf} with the stiffness
scales adapted online.
\begin{itemize}
\item \textbf{uncorrected}: no residual corrector.
\item \textbf{unconstrained}: a corrector identified without the orthogonality constraint.
\item \textbf{orthogonal (proposed)}: a corrector identified under the constraint \eqref{eq:orth_meaning}.
\end{itemize}
The basic correctors use only the
15 regressors listed in Table~\ref{tab:settings} and have 30 parameters, two per
regressor for the two outputs. Both use the
same ridge coefficient and the same identification innovations, and differ only in the
constraint. To examine how the effect of the constraint depends on the number of
parameters, the regressors are extended with random Fourier features
\citep{rahimi2007random}. These are cosines of random linear projections of the 15
standardized basic regressors and are appended to them without further standardization.
They add nonlinear regressors in any number while the corrector stays linear in its
parameters, so the same constraint applies at every size. The four extended sizes use 50 to 400 random
Fourier features, corresponding to 130 to 830 parameters, with the feature settings
detailed in Appendix~\ref{app:settings}. For each extended size, ten random feature
draws are generated, with each draw shared by both correctors for identification and
evaluation. Tables and summary figures report the means over the draws. Across the draws the sideslip error varies by at most
$0.012^{\circ}$ and the response by at most $0.12$.
With the eleven identification segments, the constraint
leaves 8 of the 30 parameters of the basic orthogonal corrector free and 808 of the 830 parameters
of the largest corrector.

Sideslip accuracy is the root-mean-square error (RMSE) of the estimated sideslip angle
against the CarMaker reference, computed over the samples with speed above $5$~m/s.
We report the mean of the run RMSEs over the twelve constant-stiffness maneuver runs,
the three maneuvers at each of the scales $0.8$, $0.9$, $1.0$, and $1.1$.
Adaptation preservation
is measured by the \emph{stiffness response}. For each axle, let $s(\lambda_{Ky})$
denote the settled stiffness scale, averaged over the final third of the run. The
estimated relative change is $s(\lambda_{Ky})/s(1)-1$, with both values obtained using
the same estimator and maneuver. Relative changes are used because each estimator's
absolute level is an effective stiffness along its own state trajectory, as
Section~\ref{sec:lsview} showed.
The nominal stiffnesses $C_{f0}$ and $C_{r0}$ are the effective-stiffness regression on
the run at nominal scale, and the reference change of a scaled run is the change of its
regression relative to that run. Because the reference is this effective change and
not the tire factor, the same $\lambda_{Ky}$ produces reference changes of different
size on the two axles. For each axle, the response is
the slope of a line through the origin fitted to the estimated changes against the
reference changes of the nine runs at a scale other than $1.0$.
A response of $1$ means that the estimate moves by the full effective change, and $0$
means that it does not move. Six of these runs are identification runs, so the response is not
an independent evaluation.

\subsection{Accuracy and stiffness response}\label{sec:sim_stealing}

\begin{figure}[!htbp]
\centering
\includegraphics[width=0.95\textwidth]{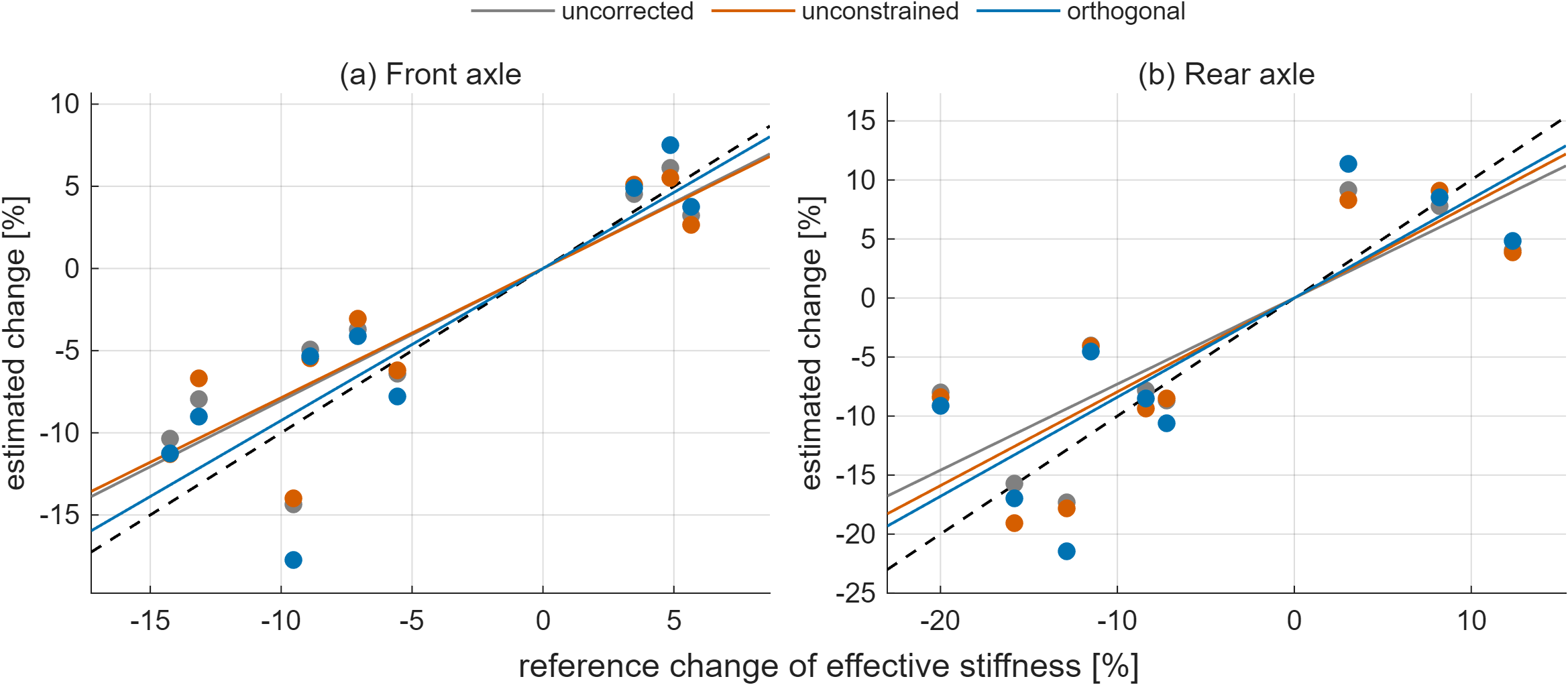}
\caption{Stiffness response on the scaled runs, basic correctors (30 parameters). One marker per
run. Dashed line: full response. Thin lines: slopes through the origin, reported in
Table~\ref{tab:cm}. (a) Front axle. (b) Rear axle.}
\label{fig:response}
\end{figure}

\begin{figure}[!htbp]
\centering
\includegraphics[width=0.95\textwidth]{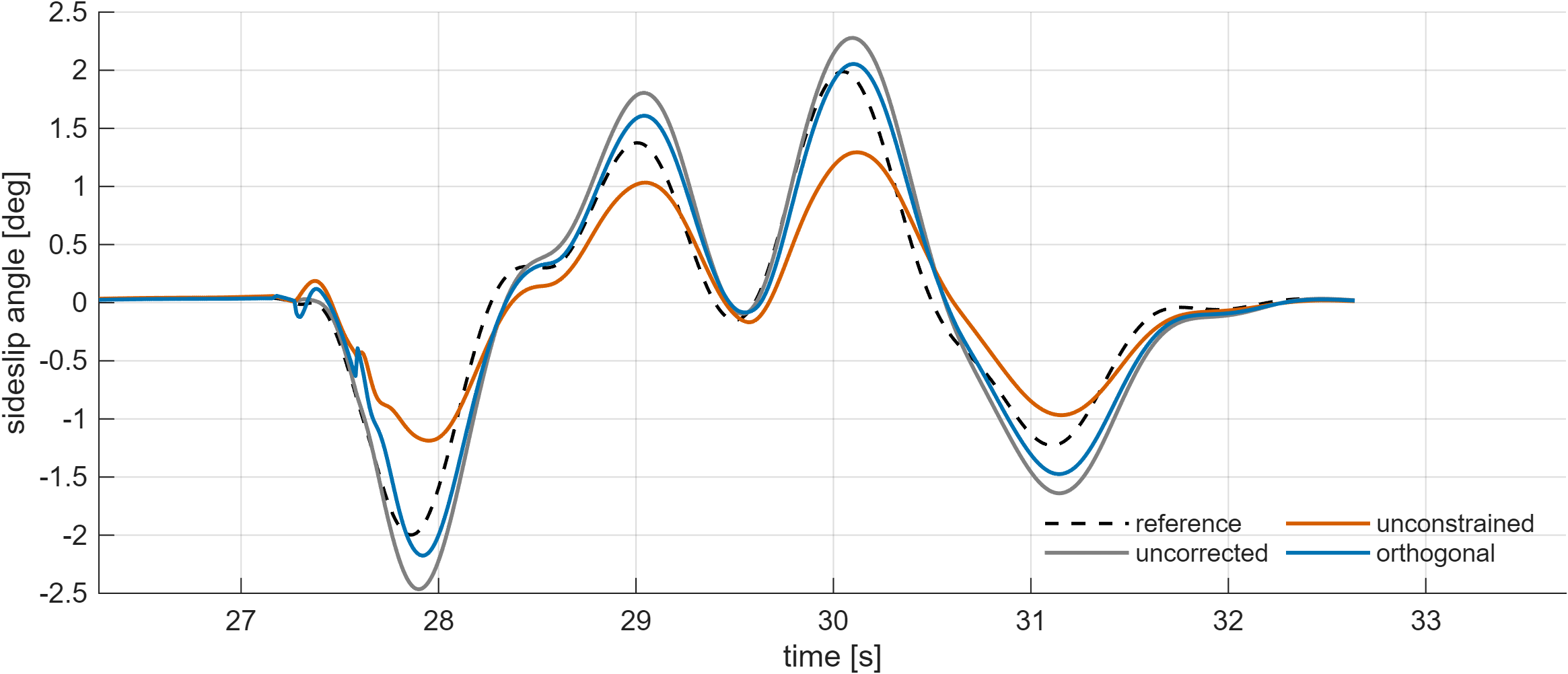}
\caption{Sideslip estimate on a double lane change at scale $0.9$, a run not used for
identification. Basic correctors (30 parameters).}
\label{fig:cm_sideslip}
\end{figure}

At constant stiffness, we compare the two metrics across corrector sizes and ask whether
added parameters improve sideslip accuracy without suppressing the stiffness adaptation.
Figure~\ref{fig:response} shows the stiffness response of the basic correctors,
with the values reported in Table~\ref{tab:cm}. The unconstrained corrector keeps the uncorrected filter's response but has a larger
sideslip error. The orthogonal corrector of the same size reduces the
sideslip RMSE by $12.7\%$ and responds to the stiffness change more fully than the
uncorrected filter on both axles. The runs at scale $0.9$, the
stiffness not used for identification, give the same ordering, with RMSEs of
$0.250^{\circ}$, $0.282^{\circ}$, and $0.219^{\circ}$ for the uncorrected filter, the
unconstrained, and the orthogonal corrector. Figure~\ref{fig:cm_sideslip} shows the
sideslip estimates on one of these runs.

\begin{figure}[!htbp]
\centering
\includegraphics[width=0.95\textwidth]{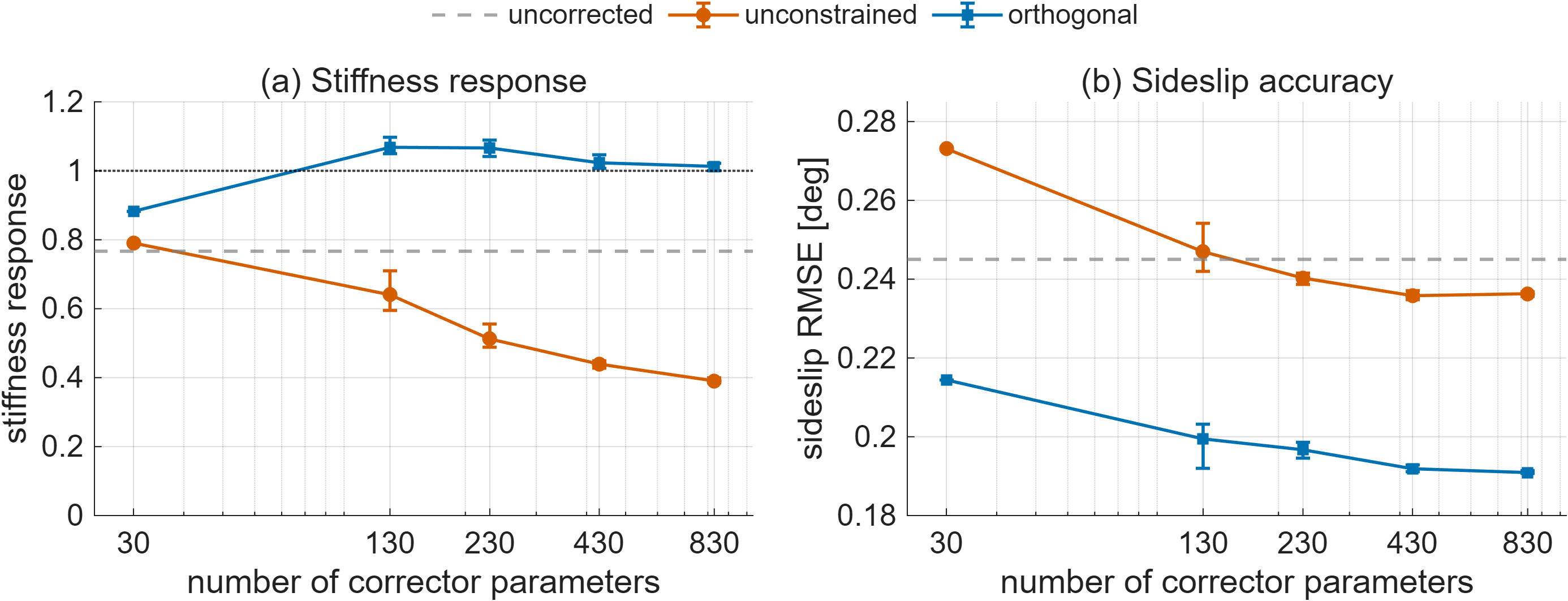}
\caption{Simulation results against the number of corrector parameters. (a) Stiffness
response, front/rear mean over the scaled runs. (b) Sideslip RMSE over the
constant-stiffness runs. Markers: mean over the feature draws. Bars: minimum to
maximum over the draws. Dashed: uncorrected filter. Dotted: full response.}
\label{fig:capacity}
\end{figure}

\FloatBarrier
If the corrector suppresses the response by absorbing innovation aligned with the
stiffness sensitivities, the suppression is expected to grow with its ability to fit
the innovation. Figure~\ref{fig:capacity} shows the two metrics as the number of corrector
parameters increases from $30$ to $830$, with the spread over the draws, and
Table~\ref{tab:cm} lists the values. The unconstrained response, averaged over the front
and rear axles, falls monotonically from $0.79$ at $30$ parameters to $0.39$ at $830$,
compared with $0.77$ for the uncorrected filter, while its sideslip error comes at most
$3.7\%$ below the uncorrected filter's. The orthogonal response stays at or above the
uncorrected filter's at every size, and its sideslip error is $12.7\%$ to $22.0\%$ below
the uncorrected filter's. The responses computed from the three runs at
scale $0.9$ alone, which are not used for identification, show the same pattern at every
size. At $430$ parameters they are $0.38$ on the front axle and $0.40$ on the rear for the
unconstrained corrector, $0.90$ and $0.92$ for the orthogonal corrector, and $0.66$ and
$0.69$ for the uncorrected filter. The constraint preserves the response as the
number of parameters grows, and the orthogonal corrector has the lowest sideslip error at
every size.

\begin{table}[!htbp]
\centering
\caption{Simulation results for all corrector sizes. Sideslip RMSE over the
constant-stiffness runs, nine of which are identification runs, and stiffness response
over the scaled runs. Rows with random Fourier features are means over the feature
draws.}
\label{tab:cm}
\footnotesize
\renewcommand{\arraystretch}{0.92}
\begin{tabular}{@{}rlccc@{}}
\toprule
Parameters & Estimator & $\beta$ RMSE [deg] & \multicolumn{2}{c}{stiffness response} \\
& & & front & rear \\
\midrule
-- & uncorrected & 0.245 & 0.80 & 0.73 \\
\midrule
30 & unconstrained & 0.273 & 0.79 & 0.79 \\
30 & orthogonal (proposed) & 0.214 & 0.92 & 0.84 \\
\addlinespace[3pt]
130 & unconstrained & 0.247 & 0.65 & 0.64 \\
130 & orthogonal (proposed) & 0.199 & 1.10 & 1.03 \\
\addlinespace[3pt]
230 & unconstrained & 0.240 & 0.52 & 0.50 \\
230 & orthogonal (proposed) & 0.197 & 1.12 & 1.01 \\
\addlinespace[3pt]
430 & unconstrained & 0.236 & 0.45 & 0.43 \\
430 & orthogonal (proposed) & 0.192 & 1.08 & 0.97 \\
\addlinespace[3pt]
830 & unconstrained & 0.236 & 0.40 & 0.38 \\
830 & orthogonal (proposed) & 0.191 & 1.07 & 0.95 \\
\bottomrule
\end{tabular}
\end{table}

\FloatBarrier
\subsection{Independent axle stiffness changes}\label{sec:independent_axles}

\begin{figure}[t]
\centering
\includegraphics[width=0.9\textwidth]{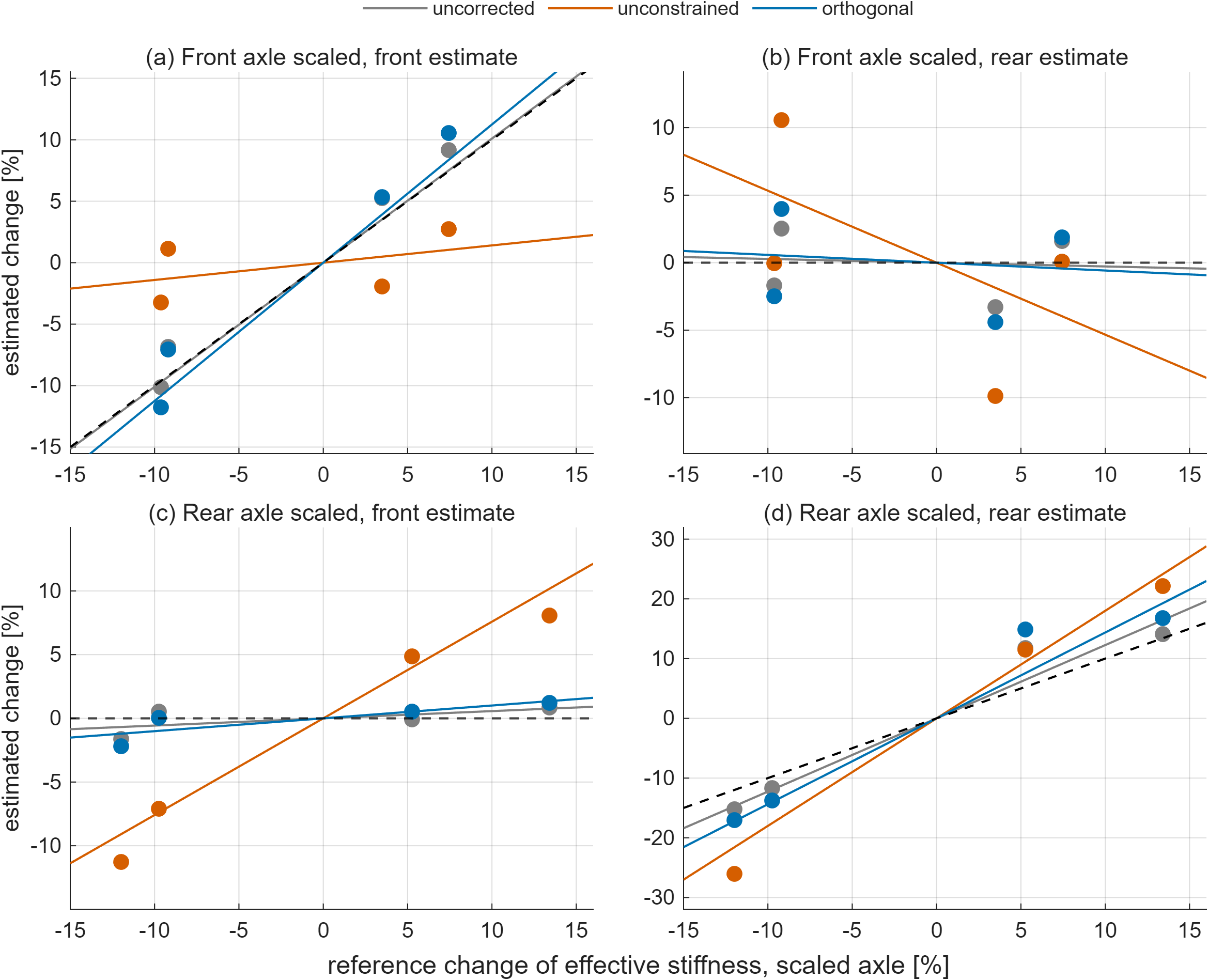}
\caption{One-axle stiffness changes at scales $0.85$ and $1.15$, basic
correctors (30 parameters). One marker per run. Rows: front axle scaled (a, b) and rear axle scaled (c, d). Columns:
estimated change of the front axle (a, c) and of the rear axle (b, d), against the
reference change of the scaled axle. Dashed lines: full response in (a, d), zero in (b,
c). Thin lines: slopes through the origin.}
\label{fig:axles}
\end{figure}

The scaled runs of Section~\ref{sec:sim_setup} apply the same stiffness factor to
both axles. The eight one-axle runs of Table~\ref{tab:matrix} set the stiffness factor to
$0.85$ or $1.15$ on one axle, with the other kept at $1.0$, for sinusoidal steering and
slalom. The one-axle runs test whether the estimators can distinguish front- and
rear-axle stiffness changes. Figure~\ref{fig:axles} shows the responses to one-axle
stiffness changes, with one row per scaled axle and one column per estimated axle. In
Figure~\ref{fig:axles}(a,d) the estimated change of the scaled axle is plotted against
its reference change, and in Figure~\ref{fig:axles}(b,c) the estimated change of the
other axle against the same reference. Although the stiffness factor of the other axle is
unchanged, its effective stiffness also changes, by $-3.6\%$ to $5.7\%$.

When the front axle changes, the uncorrected filter and the orthogonal corrector respond
on the front axle with slopes of $1.01$ and $1.12$ and leave the rear estimate nearly
unchanged. The unconstrained corrector responds with only $0.14$ on the front axle and
moves the rear estimate in the opposite direction, with $-0.53$. When the rear axle
changes, all three estimators over-respond on the rear, and the unconstrained corrector
assigns $0.76$ of the change to the front axle, compared with at most $0.10$ for the
other two. The orthogonal corrector thus keeps the axle separation of the uncorrected filter.
On these stiffness combinations absent from identification, it also
lowers the mean sideslip RMSE over the one-axle runs from $0.293^{\circ}$ to
$0.263^{\circ}$, while the unconstrained corrector raises it to $0.329^{\circ}$.

\subsection{Stiffness change during a drive}\label{sec:continuous}

The constant-stiffness runs show where each estimator settles. Here the stiffness
changes while the vehicle is driving, and the basic correctors are compared. The
cornering-stiffness scaling factor of the tire
model is switched during a $360$~s drive with repeated step steering at
constant speed. All four tires change from $1.0$ to $0.8$ at $120$~s and back at
$240$~s. A drive with the same steering and speed but without the stiffness change
provides the normalization. Each estimator's stiffness is reported relative to its own value on this
unchanged drive. The unchanged drive is part of the identification set, and the drive
with the stiffness change is not. Sideslip RMSE is computed over the first $20$~s after
the change and over the rest of the reduced-stiffness interval, from $140$~s to $240$~s.
The settled change on this drive is the mean relative change over the last $40$~s of the
reduced-stiffness interval, from $200$~s to $240$~s. Table~\ref{tab:transient} reports
it as a percentage of the effective stiffness change over the same interval, so that
$100\%$ means the estimate settled exactly at the effective change, and values above
$100\%$ mean an over-response.

\begin{figure}[!tb]
\centering
\includegraphics[width=0.95\textwidth]{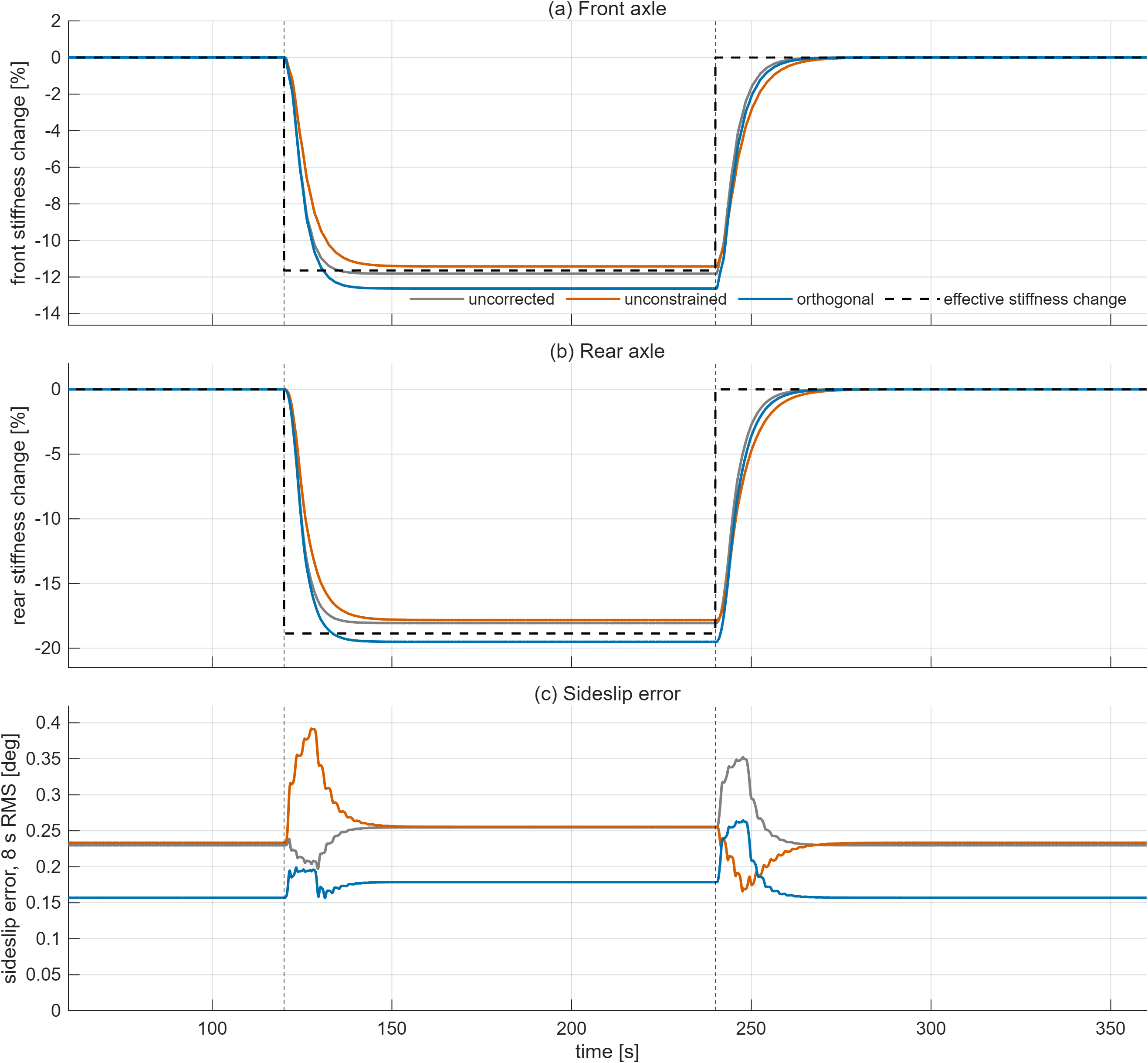}
\caption{Stiffness change during the drive at $80$~km/h, all tires to $0.8$ of nominal
from $120$~s to $240$~s (dashed lines). (a), (b) Stiffness estimate relative to the
unchanged drive, $4$~s moving average, with the effective stiffness change.
(c) Sideslip error, $8$~s moving root-mean-square value. Basic correctors (30 parameters).}
\label{fig:cm_ts}
\end{figure}

Panels (a) and (b) of Figure~\ref{fig:cm_ts} show the front and rear stiffness
estimates, and Table~\ref{tab:transient} gives the transient metrics, computed on the
unsmoothed estimates. All three estimators settle within about $20$~s, the orthogonal
corrector slightly beyond the effective change and the unconstrained corrector slightly
short of it. The unconstrained corrector reaches $90\%$ of the effective change $4$ to
$5$~s later than the other two. The slower response is consistent
with the innovation competition described in Section~\ref{sec:stealing}. In the first $20$~s after the change, the
unconstrained corrector gives the least accurate sideslip estimate of the three, and from $140$~s on it is as
accurate as the uncorrected filter. The orthogonal corrector is the most accurate in both
intervals. Figure~\ref{fig:cm_ts}(c) shows its sideslip error settling within the same
$20$~s.

\begin{table}[!htb]
\centering
\caption{Transient after the stiffness change at $120$~s, basic correctors (30 parameters),
unsmoothed estimates. Settled / effective: settled change in $200$--$240$~s as a
percentage of the effective stiffness change.}
\label{tab:transient}
\footnotesize
\setlength{\tabcolsep}{4pt}
\begin{tabular}{@{}lcccc@{}}
\toprule
& \multicolumn{2}{c}{time to $90\%$ of effective change [s]} & settled / effective [\%] & $\beta$ RMSE [deg] \\
Estimator & front & rear & front / rear & $120$--$140$~s / $140$--$240$~s \\
\midrule
uncorrected & 6.5 & 7.6 & 101 / 96 & 0.234 / 0.255 \\
unconstrained & 11.0 & 11.8 & 98 / 95 & 0.325 / 0.256 \\
orthogonal (proposed) & 6.5 & 7.1 & 108 / 103 & 0.183 / 0.179 \\
\bottomrule
\end{tabular}
\end{table}

\section{Real-vehicle evaluation}\label{sec:real}

The real-vehicle study examines whether the reduced stiffness response observed in
simulation also arises on measured vehicle data, and whether the orthogonal corrector
preserves it without losing sideslip accuracy.

\subsection{Dataset and setup}\label{sec:real_protocol}

We use the public KIT multi-surface driving dataset \citep{schulz2026multi}, recorded
with a rear-wheel-drive Hyundai IONIQ~5 driven manually in dynamic maneuvers at up to
$50$~km/h on dry asphalt. An optical velocity sensor provides the sideslip reference and
the yaw rate, and a rear-axle inertial unit the lateral acceleration, all at $100$~Hz.
Optical velocity and lateral acceleration are translated to the center of gravity using
the sensor positions. The nominal axle stiffnesses are calibrated from the identification drives
using quasi-steady relations between sideslip, steering, and lateral acceleration, with
the values listed in Appendix~\ref{app:settings}. The estimator settings, regressors, and ridge
coefficient are those of the simulation.

\begin{figure}[t]
\centering
\includegraphics[width=0.95\textwidth]{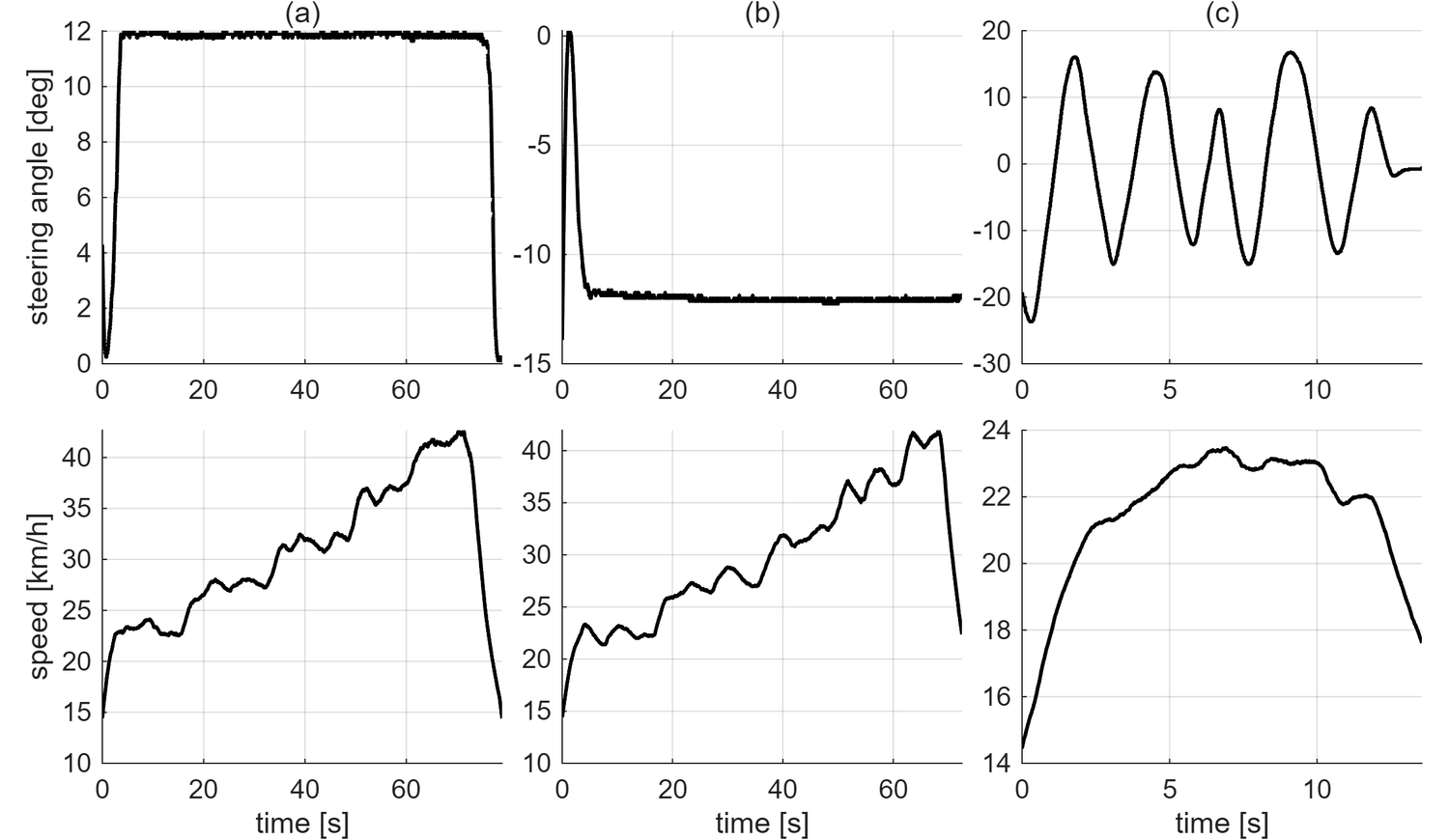}
\caption{Real-vehicle drives, steering angle (top) and speed (bottom). (a), (b)
Identification drives. (c) One evaluation drive.}
\label{fig:kit_inputs}
\end{figure}

Six drives on one asphalt track are used for identification, and four drives for
evaluation, three on the same track and one on a second asphalt track.
Figure~\ref{fig:kit_inputs} shows identification and evaluation drives.
During the identification drives the vehicle turns mostly in one direction, whereas the
single-track model is symmetric between left and right turns. Each identification drive is therefore also
used with the signs of steering angle, yaw rate, and lateral acceleration reversed, as a
separate constraint segment, $J=12$. We report the arithmetic mean of the four drive
RMSEs, each computed for $v>4$~m/s. The same procedure for repeated identification,
evaluation, and averaging over feature draws is used as in simulation.
We use the 430-parameter correctors, since increasing the size to $830$ yields little
further improvement in simulated sideslip accuracy in Table~\ref{tab:cm}.

On the vehicle the tire stiffness cannot be changed, so the mismatch is created on the
model side. Both nominal axle stiffnesses of the filter are multiplied by a factor
$\kappa\in\{0.8,\,0.9,\,1.0,\,1.1,\,1.25\}$. The vehicle is unchanged, so an estimator
that compensates the offset moves its stiffness scale to $s(1)/\kappa$, where $s(1)$ is
its settled value at $\kappa=1$. The relative change $1/\kappa-1$ that full
compensation requires therefore serves as the reference change, in place of the
effective stiffness change of Section~\ref{sec:sim_setup}. The response then
measures the fraction of the required change that the stiffness scale makes, not how
closely it tracks a known physical stiffness. At each $\kappa$ the identification is
repeated in full, and each feature draw is held fixed across the offsets. The
fixed-stiffness filter with the offset nominal stiffnesses produces the identification
innovations, the correctors are refitted on the
identification segments, and the adaptive estimators run on the four evaluation drives. The
settled stiffness scale $s(\kappa)$ of a drive is its mean over the second half of the
samples with $v>4$~m/s, and the estimated relative change is $s(\kappa)/s(1)-1$. For each drive and axle, the estimated changes at the five offsets are regressed
against the required changes through the origin. The stiffness response is the mean
of these slopes over the four evaluation drives. Sideslip RMSE is
reported at $\kappa=1$.

\subsection{Accuracy and stiffness response}\label{sec:real_results}

Table~\ref{tab:kit} gives the two metrics with $430$ parameters, averaged over the
feature draws. As in simulation, the unconstrained corrector loses most of its response
to the offsets, with $0.35$ on the front axle and $0.24$ on the rear, compared with
$0.99$ and $1.00$ for the uncorrected filter. The orthogonal corrector retains $0.88$ and $0.70$.
The sideslip errors are $0.79^{\circ}$ for the uncorrected filter, $0.79^{\circ}$ for
the unconstrained corrector, and $0.83^{\circ}$ for the orthogonal corrector. On the
vehicle, the orthogonal corrector retains the response at a slightly higher
sideslip error, and the unconstrained corrector gives up the response without a lower
sideslip error. The orthogonal corrector's response exceeds the unconstrained
corrector's on both axles for every feature draw.

\begin{table}[t]
\centering
\caption{Real-vehicle results with $430$ parameters. For the correctors, entries report
the mean over the feature draws, with the minimum and maximum in brackets.}
\label{tab:kit}
\footnotesize
\begin{tabular}{@{}lccc@{}}
\toprule
Estimator & $\beta$ RMSE [deg] & response, front & response, rear \\
\midrule
uncorrected & 0.79 & 0.99 & 1.00 \\
unconstrained & 0.79 [0.69, 0.92] & 0.35 [0.28, 0.45] & 0.24 [0.14, 0.37] \\
orthogonal (proposed) & 0.83 [0.77, 0.91] & 0.88 [0.73, 1.05] & 0.70 [0.44, 0.98] \\
\bottomrule
\end{tabular}
\end{table}

\begin{figure}[t]
\centering
\includegraphics[width=0.95\textwidth]{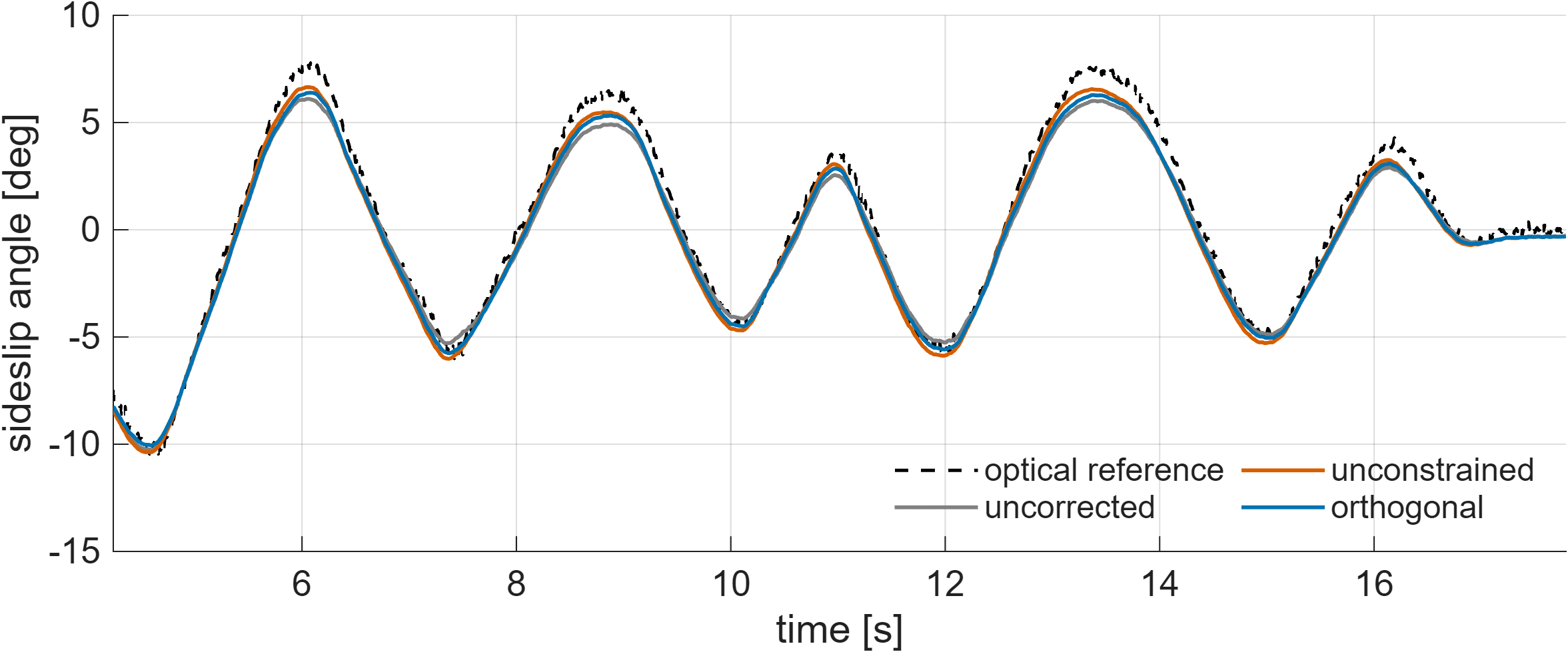}
\caption{Sideslip estimate on one evaluation drive against the optical reference.
430-parameter correctors, one feature draw.}
\label{fig:kit_ts}
\end{figure}

Figure~\ref{fig:kit_ts} shows the sideslip estimates on the evaluation drive of
Figure~\ref{fig:kit_inputs}(c). All three
estimators follow the optical reference through sideslip excursions of about
$\pm 10^{\circ}$, within about $2^{\circ}$ of the reference, and the three estimates lie within about
$1^{\circ}$ of each other.

\begin{figure}[t]
\centering
\includegraphics[width=0.95\textwidth]{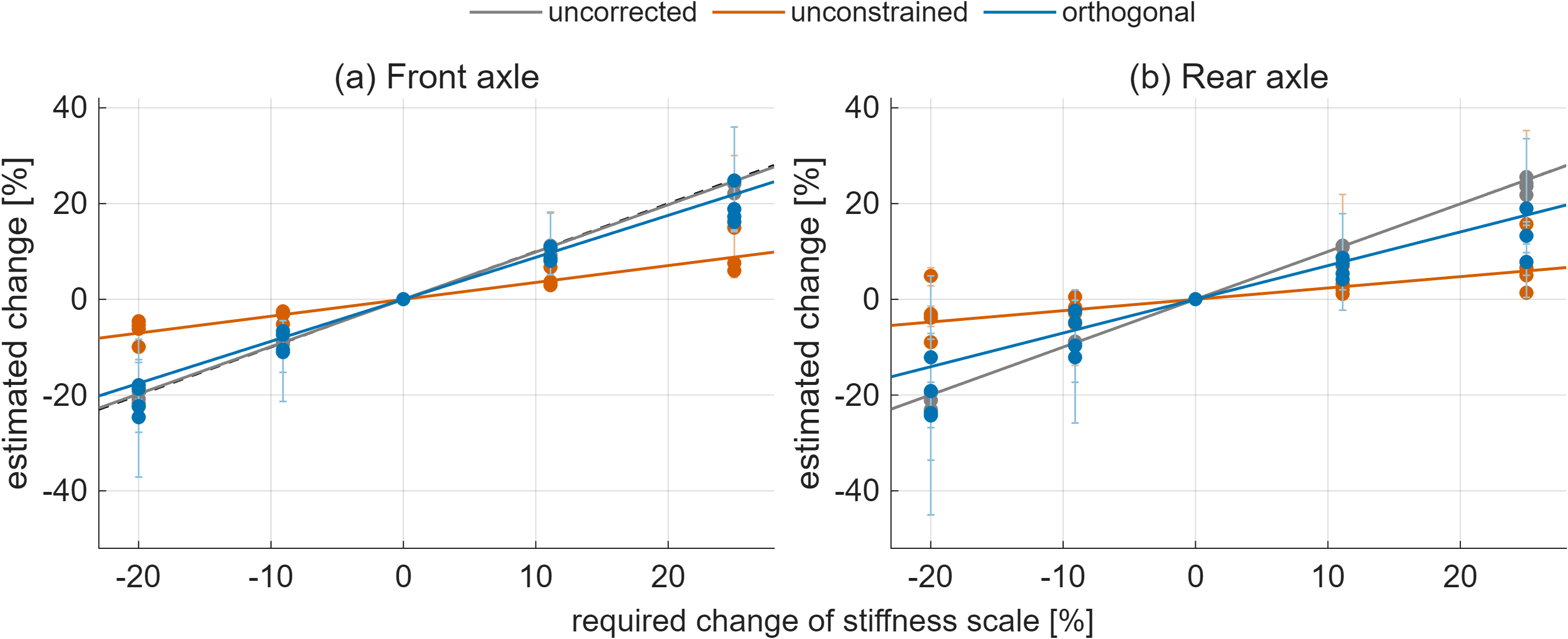}
\caption{Stiffness response on the vehicle, 430-parameter correctors. One marker per
evaluation drive and offset, mean over the feature draws, with thin bars from the
minimum to the maximum over the draws. Dashed line: full compensation. Solid lines:
slopes through the origin, reported in Table~\ref{tab:kit}. (a) Front axle. (b) Rear
axle.}
\label{fig:kit_response}
\end{figure}

Figure~\ref{fig:kit_response} shows the runs from which the response is computed, and
each marker represents the mean over the draws. The unconstrained corrector's response is at most
$0.57$ on any drive and axle, and the uncorrected filter follows the required change on
all four drives. The orthogonal corrector's response varies across drives. Its front
stiffness estimate over-responds on one drive, and its rear estimate responds less than
the front on the other three.

\begin{figure}[t]
\centering
\includegraphics[width=0.95\textwidth]{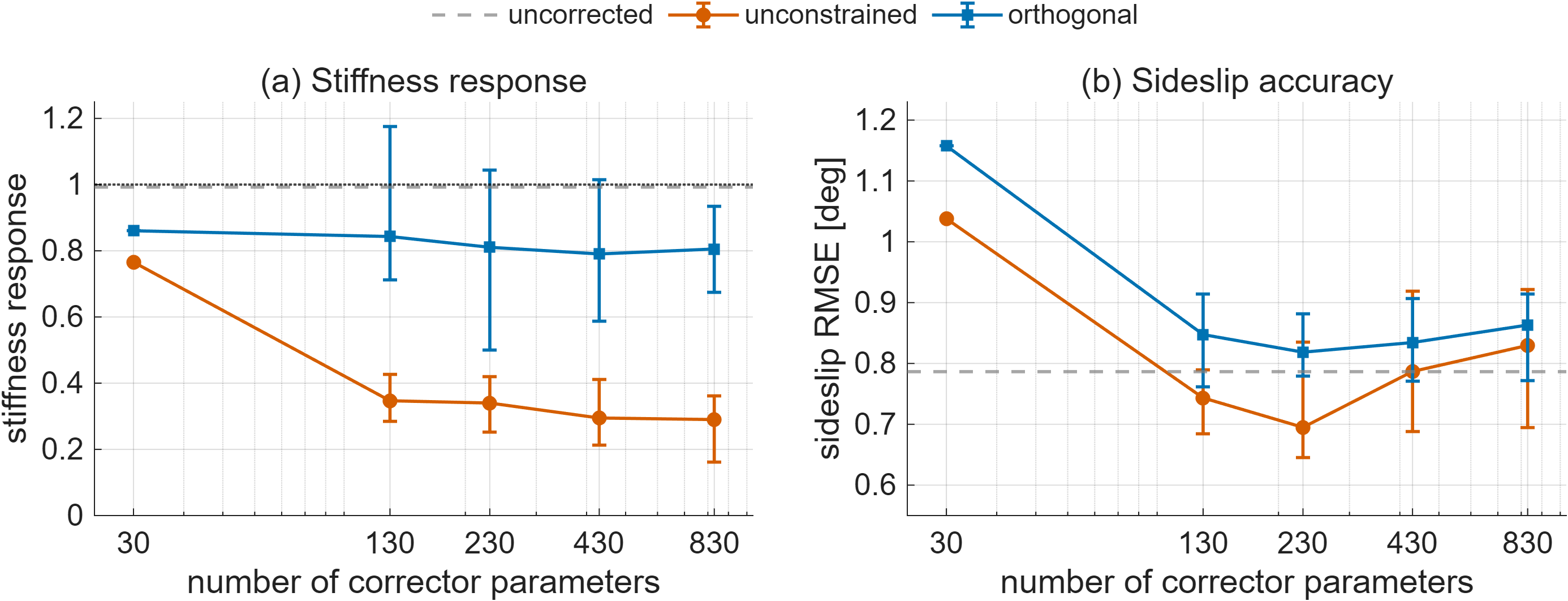}
\caption{Real-vehicle results against the number of corrector parameters. (a) Stiffness
response, front/rear mean. (b) Sideslip RMSE at $\kappa=1$. Markers: mean over the
feature draws. Bars: minimum to maximum over the draws. Dashed: uncorrected filter.
Dotted: full compensation.}
\label{fig:kit_capacity}
\end{figure}

Figure~\ref{fig:kit_capacity} shows the response and the sideslip error against the
number of corrector parameters, with the spread over the draws. The unconstrained
response, averaged over the front and rear axles, declines with size, as it does in
simulation in Figure~\ref{fig:capacity}, from $0.77$ at $30$ parameters to $0.29$ at
$830$. The orthogonal response is higher at every size, and this ordering holds for every
feature draw. For the correctors with random Fourier features, the
orthogonal response averaged over the draws ranges from $0.79$ to $0.84$, below the
uncorrected filter's $0.99$. Both correctors have a higher sideslip error than the
uncorrected filter at $30$ parameters. From $130$ parameters on, the mean sideslip
error over the draws ranges from $0.70^{\circ}$ to $0.83^{\circ}$ for the unconstrained
corrector and from $0.82^{\circ}$ to $0.86^{\circ}$ for the orthogonal corrector,
compared with $0.79^{\circ}$ for the uncorrected filter. The spread of the sideslip error
over the draws is concentrated on one evaluation drive with $73$~s of sustained
cornering.

\section{Discussion and conclusions}\label{sec:discussion}\label{sec:conclusion}

Learned residual correction can improve state accuracy while suppressing the stiffness
response of a joint estimator. The proposed sensitivity-orthogonal parameterization
leaves the sensitivity-aligned component of the innovation unchanged and preserves the
static least-squares stiffness estimate on each identification segment.

In full-vehicle simulation, the stiffness response of the unconstrained corrector
declined as its parameter count grew, whereas the orthogonal corrector responded at
least as fully as the uncorrected filter at every size, kept the uncorrected filter's
axle separation, adapted
as quickly to a stiffness change during a drive, and had the lowest sideslip error. On
the real-vehicle data, it retained most of the response to nominal-stiffness offsets that
the unconstrained corrector lost, at a slightly higher sideslip error.

The invariance of Section~\ref{sec:orth_meaning} is exact for the static least-squares
estimate on the identification segments. Model discrepancy that is itself aligned with
the stiffness sensitivities is preserved by the constraint as well, so the stiffness
scales estimate an effective stiffness of the single-track model rather than the tire
stiffness. The guarantee of the constraint holds on the segments where it is enforced
and does not extend to the extrapolation of the corrector beyond them. The
identification set should therefore cover the driving conditions in which the estimator
is to operate.

The constraint acts on the innovation that excitation has already created, so the
excitation-based scheduling of adaptation reviewed in Section~\ref{sec:related} is
complementary to it. The same orthogonality constraint can be constructed using
sensitivities computed from a nonlinear axle-force model. Learned correction and online
parameter adaptation can coexist in one filter when the corrector cannot reproduce the
innovation that a parameter change produces.

\section*{Disclosure statement}
The authors report there are no competing interests to declare.

\section*{Declaration of generative AI use}
During the preparation of this work, the authors used Claude Opus 5.5 (Anthropic) to
improve the language and to assist with the analysis scripts. The authors reviewed and
edited all outputs and take full responsibility for the content of this article.

\section*{Data availability statement}
The real-vehicle data are publicly available in the KIT multi-surface driving dataset
at \url{https://doi.org/10.35097/44a91t97pmnha1k9} \citep{schulz2026multi}. The
simulation data were generated with the commercial software IPG CarMaker and are
available from the corresponding author upon reasonable request.

\section*{Funding}
This work was supported by the National Research Foundation of Korea (NRF) grant funded
by the Korea government (MSIT) (No.~RS-2026-25473397). This work was supported by the
Korea Planning \& Evaluation Institute of Industrial Technology (KEIT) grant funded by the
Korean Ministry of Trade, Industry and Resources (MOTIR) (RS-2026-25544705, Development of
a Scalable Heterogeneous Zonal Chassis SW Platform and Vehicle-Adaptive Drive Modules for
Global OEM Customization).

\bibliographystyle{tfp}
\bibliography{ref}

\FloatBarrier
\appendix
\section{Estimator and identification settings}\label{app:settings}

Table~\ref{tab:settings} lists the vehicle parameters, the filter and identification
settings used throughout, and the regressors of the basic corrector.

\begin{table}[h]
\centering
\caption{Vehicle parameters, estimator and identification settings, and the 15 regressors
of the basic corrector. Vehicle parameters are from the CarMaker datasheet (simulation)
and the dataset description (real vehicle). Process-noise entries are per sample at
$\Delta t = 0.01$~s. Superscript m denotes a measured signal.}
\label{tab:settings}
\footnotesize
\begin{tabular}{@{}p{0.27\textwidth} p{0.33\textwidth} p{0.29\textwidth}@{}}
\toprule
Quantity & Simulation (CarMaker) & Real vehicle (IONIQ~5) \\
\midrule
$l_f$, $l_r$ [m] & $0.97$, $1.57$ & $1.563$, $1.437$ \\
$m$ [kg], $I_z$ [kg\,m$^2$] & $1598$, $1800$ & $2254$, $2193$ \\
$C_{f0}$, $C_{r0}$ [kN/rad] & $89.9$, $91.8$ & $157.3$, $276.4$ \\
steering input [rad] & Map from the handwheel angle $\delta_{\mathrm{hw}}$ [rad] to the road-wheel angle:\newline $\delta = 0.0723\,\delta_{\mathrm{hw}} + 8.4\times10^{-5}\,\delta_{\mathrm{hw}}^{3}$ & Dataset single-track angle \\
signals & CarMaker outputs at $100$~Hz & Optical velocity sensor (Kistler Correvit S-Motion) and rear-axle inertial unit, recorded at $1$~kHz and averaged over blocks of 10 samples to $100$~Hz \\
no stiffness update below & $v < 5$~m/s & $v < 3$~m/s \\
evaluation window & $v > 5$~m/s & $v > 4$~m/s \\
effective-stiffness regression & through the origin, per axle; samples with $v > 5$~m/s and $\lvert\alpha\rvert < 0.75^{\circ}$; on the drive with a stiffness change, last $40$~s of each interval & -- \\
\midrule
\multicolumn{3}{@{}l}{Shared filter settings} \\
$R$ & \multicolumn{2}{l}{$\mathrm{diag}\big((0.15^{\circ}/\mathrm{s})^2,\ (0.05~\mathrm{m/s^2})^2\big)$} \\
$Q$, dynamic states $\beta, r$ & \multicolumn{2}{l}{$10^{-8}$, $10^{-8}$} \\
$Q$, stiffness scales $s_f, s_r$ & \multicolumn{2}{l}{$(3\times 10^{-4})^2$ each; $0$ in the fixed-stiffness filter} \\
$Q$, yaw-rate bias $b_r$ & \multicolumn{2}{l}{$(2\times 10^{-5}~\mathrm{rad/s})^2$} \\
stiffness-scale bounds & \multicolumn{2}{p{0.62\textwidth}@{}}{$[0.3,\ 2.5]$, enforced by a covariance-weighted projection of the joint state estimate with the corresponding covariance update} \\
\midrule
\multicolumn{3}{@{}l}{Shared identification settings} \\
ridge coefficient $\rho$ & \multicolumn{2}{l}{$10^{-3}$} \\
constraint segments & \multicolumn{2}{p{0.62\textwidth}@{}}{One per run: $J=11$ (simulation), $12$ (real vehicle: six drives and their reflections)} \\
rank of $\mathcal{C}$ & $22$ & $18$ for the basic corrector (twelve of the fifteen regressors change sign under the reflection), $24$ with random Fourier features \\
random Fourier features & \multicolumn{2}{p{0.62\textwidth}@{}}{$\cos(\Omega^{\top}\tilde{\phi} + \mathbf{b})$, appended to $\tilde{\phi}$, the vector of the 15 standardized basic regressors below. $\Omega$ has 15 rows and one column per added cosine term, $\Omega_{ij} \sim \mathcal{N}(0, 1/15)$; $\mathbf{b}$ holds one phase per term, $b_j \sim \mathcal{U}(0, 2\pi)$} \\\midrule
\multicolumn{3}{@{}l}{Regressors of the basic corrector} \\
constant & \multicolumn{2}{l}{$1$} \\
steering & \multicolumn{2}{l}{$\delta_k,\ \delta_{k-1},\ \delta_{k-2}$} \\
delayed measurements & \multicolumn{2}{l}{$r^{\mathrm{m}}_{k-1},\ r^{\mathrm{m}}_{k-2},\ a^{\mathrm{m}}_{y,k-1},\ a^{\mathrm{m}}_{y,k-2},\ v_{k-1}$} \\
odd nonlinearities & \multicolumn{2}{l}{$a^{\mathrm{m}}_{y,k-1}\lvert a^{\mathrm{m}}_{y,k-1}\rvert,\ \delta_k\lvert\delta_k\rvert,\ r^{\mathrm{m}}_{k-1}\lvert r^{\mathrm{m}}_{k-1}\rvert,\ (a^{\mathrm{m}}_{y,k-1})^{3}$} \\
products & \multicolumn{2}{l}{$\delta_k\,a^{\mathrm{m}}_{y,k-1},\ \delta_k\,v_{k-1}$} \\
\bottomrule
\end{tabular}
\end{table}

\end{document}